%% file: arxiv.tex
\documentclass[10pt,twocolumn,letterpaper]{article}

\usepackage[pagenumbers]{cvpr} 

\usepackage{graphicx}
\usepackage{amsmath}
\usepackage{amssymb}
\usepackage{booktabs}

\makeatletter
\@namedef{ver@everyshi.sty}{}
\makeatother
\usepackage[hashEnumerators,hybrid]{markdown}
\usepackage{xcolor}
\usepackage[utf8x]{inputenc}
\usepackage{pgf}
\usepackage{colortbl}
\usepackage{tikz}
\usepackage{tabularx}
\usepackage{subcaption}
\usepackage{placeins}
\usepackage{float}

\usepackage[pagebackref,breaklinks,colorlinks]{hyperref}

\usepackage[capitalize]{cleveref}
\crefname{section}{Sec.}{Secs.}
\Crefname{section}{Section}{Sections}
\Crefname{table}{Table}{Tables}
\crefname{table}{Tab.}{Tabs.}

\newcommand{\spm}[1]{{\scriptsize$\pm$#1}}
\newcommand*{\cm}{\checkmark}

\DeclareMathOperator*{\argmax}{arg\,max}

\newcolumntype{Y}{>{\centering\arraybackslash}X}

\def\cvprPaperID{3423} 
\def\confName{CVPR}
\def\confYear{2022}

\begin{document}

\title{DAFormer: Improving Network Architectures and Training Strategies for Domain-Adaptive Semantic Segmentation}

\author{Lukas Hoyer\\
ETH Zurich\\
{\tt\small lhoyer@vision.ee.ethz.ch}
\and
Dengxin Dai\\
MPI for Informatics\\
{\tt\small ddai@mpi-inf.mpg.de}
\and
Luc Van Gool\\
ETH Zurich \& KU Leuven\\
{\tt\small vangool@vision.ee.ethz.ch}
}
\maketitle

\input{sec_abstract}

\input{sec_intro}
\input{sec_related_work}
\input{sec_methods}

\input{sec_experiments}

\input{sec_conclusions}

\clearpage

{\small
\bibliographystyle{ieee_fullname}
\bibliography{egbib}
}

\clearpage

\noindent\textbf{\Large Supplementary Material}

\makeatletter
\renewcommand{\theHsection}{papersection.\number\value{section}} 
\renewcommand{\thesection}{\Alph{section}}
\renewcommand{\thefigure}{S\arabic{figure}}
\renewcommand{\thetable}{S\arabic{table}}
\setcounter{section}{0}

\setcounter{figure}{0}
\setcounter{table}{0}
\makeatother

\input{sec_supplement}

\end{document}

%% file: sec_abstract.tex
\begin{abstract}
As acquiring pixel-wise annotations of real-world images for semantic segmentation is a costly process, a model can instead be trained with more accessible synthetic data and adapted to real images without requiring their annotations. This process is studied in unsupervised domain adaptation (UDA). Even though a large number of methods propose new adaptation strategies, they are mostly based on outdated network architectures. 
As the influence of recent network architectures has not been systematically studied, we first benchmark different network architectures for UDA and newly reveal the potential of Transformers for UDA semantic segmentation. Based on the findings, we propose a novel UDA method, DAFormer. 
The network architecture of DAFormer consists of a Transformer encoder and a multi-level context-aware feature fusion decoder. It is enabled by three simple but crucial training strategies to stabilize the training and to avoid overfitting to the source domain:
While (1) Rare Class Sampling on the source domain improves the quality of the pseudo-labels by mitigating the confirmation bias of self-training toward common classes, (2) a Thing-Class ImageNet Feature Distance and (3) a learning rate warmup promote feature transfer from ImageNet pretraining.
DAFormer represents a major advance in UDA. It improves the state of the art by 10.8 mIoU for GTA$\rightarrow$Cityscapes and 5.4 mIoU for Synthia$\rightarrow$Cityscapes and enables learning even difficult classes such as train, bus, and truck well. The implementation is available at
{\footnotesize\url{https://github.com/lhoyer/DAFormer}}.
\end{abstract}

%% file: sec_intro.tex
\section{Introduction}
\label{sec:introduction}

\begin{figure}
    \centering
    \includegraphics[width=0.95\linewidth]{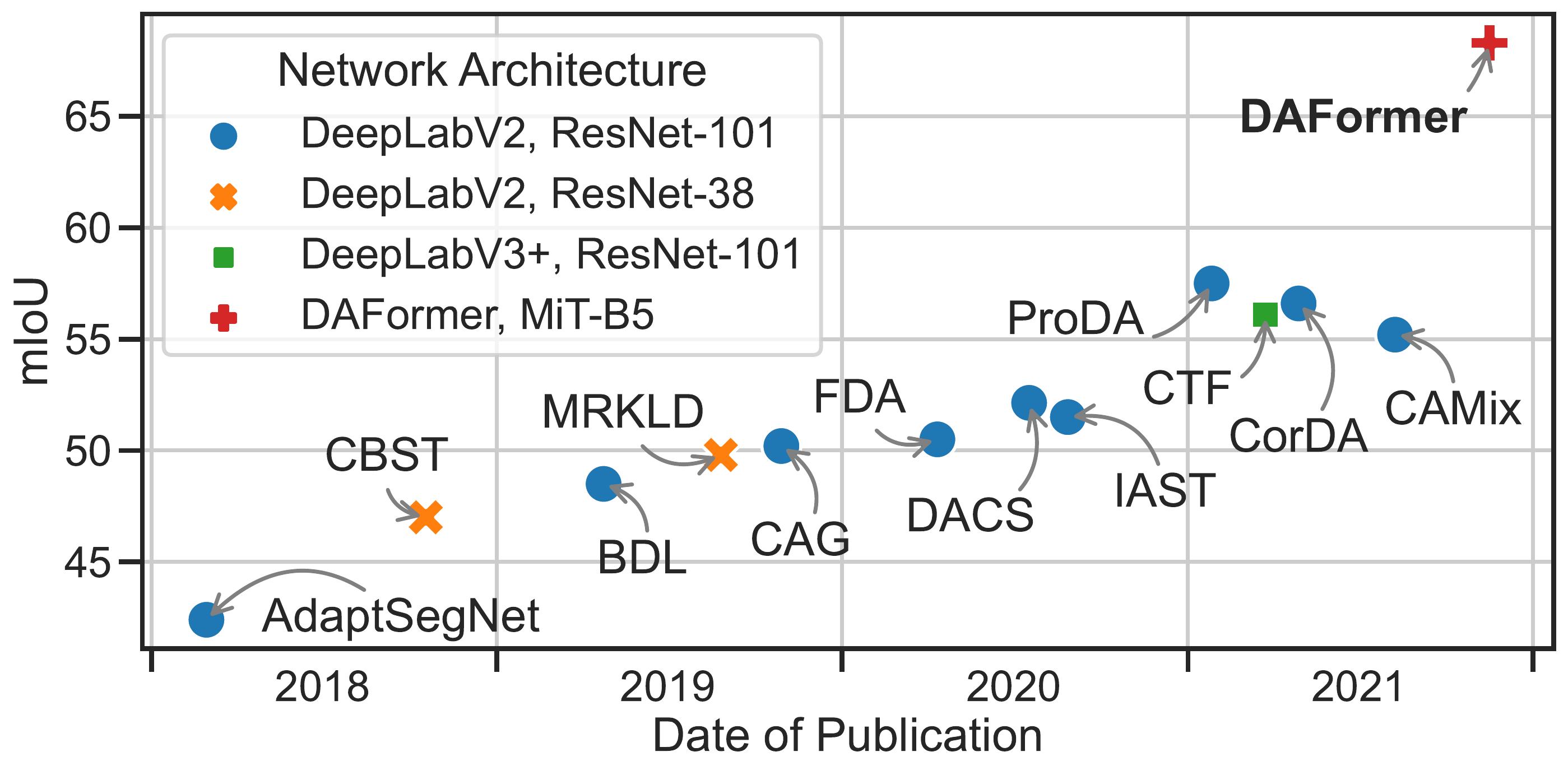}
    \vspace{-0.3cm}
    \caption{Progress of UDA over time on GTA$\rightarrow$Cityscapes. Most previous UDA methods are evaluated using the outdated DeepLabV2 architecture. We rethink the design of the network architecture as well as its training strategies for UDA and propose DAFormer, significantly outperforming previous methods.}
    \label{fig:uda_over_time}
\end{figure}

In the last few years, neural networks have achieved overwhelming performance on many computer vision tasks. However, they require a large amount of annotated data in order to be trained properly. For semantic segmentation, annotations are particularly costly as every pixel has to be labeled. For instance, it takes 1.5 hours to annotate a single image of Cityscapes~\cite{cordts2016cityscapes} while, for adverse weather conditions, it is even 3.3 hours~\cite{sarkadis2021acdc}. One idea to circumvent this issue is training with synthetic data~\cite{ros2016synthia, richter2016playing}. However, commonly used CNNs~\cite{lecun1998gradient} are sensitive to domain shift and generalize poorly from synthetic to real data. This issue is addressed in unsupervised domain adaptation (UDA) by adapting the network trained with source (synthetic) data to target (real) data without access to target labels. 

Previous UDA methods mostly evaluated their contributions using a DeepLabV2~\cite{chen2017deeplab} or FCN8s~\cite{long2015fully} network architecture with ResNet~\cite{he2016deep} or VGG~\cite{simonyan2014very} backbone in order to be comparable to previously published works. However, even their strongest architecture (DeepLabV2+ResNet101) is outdated in the area of supervised semantic segmentation. For instance, it only achieves a supervised performance of 65 mIoU~\cite{tsai2018learning} on Cityscapes while recent networks reach up to 85 mIoU~\cite{yuan2020object, tao2020hierarchical}.
Due to the large performance gap, it stands to question whether using outdated network architectures can limit the overall performance of UDA and can also misguide the benchmark progress of UDA. In order to answer this question, this work studies the influence of the network architecture for UDA, compiles a more sophisticated architecture, and successfully applies it to UDA with a few simple, yet crucial training strategies.
Naively using a more powerful network architecture for UDA might be suboptimal as it can be more prone to overfitting to the source domain.
Based on a study of different semantic segmentation architectures evaluated in a UDA setting, we compile DAFormer, a network architecture tailored for UDA (Sec~\ref{sec:methods_architecture}). It is based on recent Transformers~\cite{dosovitskiy2020image, xie2021segformer}, which have been shown to be more robust than the predominant CNNs~\cite{bhojanapalli2021understanding}. We combine them with a context-aware multi-level feature fusion, which further enhances the UDA performance. 
To the best of our knowledge, DAFormer is the first work to reveal the significant potential of Transformers for UDA semantic segmentation.

Since more complex and capable architectures are more prone to adaptation instability and overfitting to the source domain, in this work,
we introduce three training strategies to UDA to address these issues (Sec.~\ref{sec:methods_training_strategies}).
First, we propose Rare Class Sampling (RCS) to consider the long-tail distribution of the source domain, which hinders the learning of rare classes, especially in UDA due to the confirmation bias of self-training toward common classes. By frequently sampling images with rare classes, 
the network can learn them more stably, 
which improves the quality of pseudo-labels and reduces the confirmation bias.
Second, we propose a Thing-Class ImageNet Feature Distance (FD), which distills knowledge from diverse and expressive ImageNet features in order to regularize the source training. This is particularly helpful as the source domain is limited to only a few instances of certain classes (low diversity), which have a different appearance than the target domain (domain shift). Without FD this would result in learning less expressive and source-domain-specific features. As ImageNet features were trained for thing-classes, we restrict the FD to regions of the image that are labeled as a thing-class.
And third, we introduce learning rate warmup~\cite{goyal2017accurate} newly to UDA. By linearly increasing the learning rate up to the intended value in the early training, the learning process is stabilized and features from ImageNet pretraining can be better transferred to semantic segmentation.

DAFormer outperforms previous methods by a large margin (see Fig.~\ref{fig:uda_over_time}) supporting our hypothesis that the network architecture and appropriate training strategies play an important role for UDA. On GTA$\rightarrow$Cityscapes, we improve the mIoU from 57.5~\cite{zhang2021prototypical} to 68.3 and on Synthia$\rightarrow$Cityscapes from 55.5~\cite{zhang2021prototypical} to 60.9. In particular, DAFormer learns even difficult classes that previous methods struggled with. For instance, we improve the class \emph{train} from 16 to 65 IoU, \emph{truck} from 49 to 75 IoU, and \emph{bus} from 59 to 78 IoU on GTA$\rightarrow$Cityscapes.
Overall, DAFormer represents a major advance in UDA.
Our framework can be trained in one stage on a single consumer RTX~2080~Ti GPU within 16 hours, which simplifies its usage compared to previous methods such as ProDA~\cite{zhang2021prototypical}, which requires training multiple stages on four V100 GPUs for several days.

%% file: sec_related_work.tex
\section{Related Work}
\label{sec:related_work}

\paragraph{Semantic Image Segmentation}

Since the introduction of Convolutional Neural Networks (CNNs)~\cite{lecun1998gradient} for semantic segmentation by Long \etal~\cite{long2015fully}, they have been dominating the field. Typically, semantic segmentation networks follow an encoder-decoder design~\cite{long2015fully, badrinarayanan2017segnet, ronneberger2015u}. To overcome the problem of the low spatial resolution at the bottleneck, remedies such as skip connections~\cite{ronneberger2015u}, dilated convolutions~\cite{yu2015multi, chen2014semantic}, or resolution preserving architectures~\cite{sun2019high} were proposed. Further improvements were achieved by harnessing context information, for instance using pyramid pooling~\cite{zhao2017pyramid, chen2017deeplab, chen2018encoder, hoyer2019grid} or attention modules~\cite{wang2018non, fu2019dual, huang2019interlaced, yuan2020object}.
Inspired by the success of the attention-based Transformers~\cite{vaswani2017attention} in natural language processing, they were adapted to image classification~\cite{dosovitskiy2020image, touvron2021training} and semantic segmentation~\cite{xie2021segformer, zheng2021rethinking, liu2021swin} achieving state-of-the-art results.
For image classification, CNNs were shown to be sensitive to distribution shifts such as image corruptions~\cite{hendrycks2018benchmarking}, adversarial noise~\cite{szegedy2013intriguing}, or domain shifts~\cite{hendrycks2021many}. Recent works~\cite{paul2021vision, bhojanapalli2021understanding, naseer2021intriguing} show that Transformers are more robust than CNNs with respect to these properties. While CNNs focus on textures~\cite{geirhos2018imagenet}, Transformers put more importance on the object shape~\cite{bhojanapalli2021understanding, naseer2021intriguing}, which is more similar to human vision~\cite{geirhos2018imagenet}.
For semantic segmentation, ASPP~\cite{chen2018encoder} and skip connections~\cite{ronneberger2015u}  were reported to increase the robustness~\cite{kamann2021benchmarking}. Further, Xie~\etal~\cite{xie2021segformer} showed that their Transformer-based architecture improves the robustness over CNN-based networks.
To the best of our knowledge, the influence of recent network architectures on the UDA performance of semantic segmentation has not been systematically studied yet.

\paragraph{Unsupervised Domain Adaptation (UDA)}
\label{sec:related_work_uda}

UDA methods can be grouped into adversarial training and self-training approaches. Adversarial training methods aim to align the distributions of source and target domain at input~\cite{hoffman2018cycada, gong2021dlow}, feature~\cite{hoffman2016fcns, tsai2018learning}, output~\cite{tsai2018learning, vu2019advent}, or patch level~\cite{tsai2019domain} in a GAN framework~\cite{goodfellow2014generative, ganin2016domain}. Using multiple scales~\cite{chen2018road, tsai2018learning} or category information~\cite{luo2019taking, du2019ssf, wang2020differential} for the discriminator can refine the alignment. 
In self-training, the network is trained with pseudo-labels~\cite{lee2013pseudo} for the target domain. Most of the UDA methods pre-compute the pseudo-labels offline, train the model, and repeat the process~\cite{zou2018unsupervised, zou2019confidence, yang2020fda, dai2019model}. 
Alternatively, pseudo-labels can be calculated online during the training. In order to avoid training instabilities, pseudo-label prototypes~\cite{zhang2021prototypical} or consistency regularization~\cite{tarvainen2017mean, sohn2020fixmatch} based on data augmentation~\cite{choi2019self, melaskyriazi2021pixmatch, araslanov2021self} or domain-mixup~\cite{tranheden2021dacs, zhou2021context} are used.
Several methods also combine adversarial and self-training~\cite{li2019bidirectional, wang2020classes, kim2020learning}, train with auxiliary tasks~\cite{vu2019dada, wang2021domain, hoyer2021improving}, or perform test-time UDA~\cite{wang2022continual}.

Datasets are often imbalanced and follow a long-tail distribution, which biases models toward common classes~\cite{wang2017learning}. Strategies to address this problem
are re-sampling~\cite{he2008adasyn, wei2021crest, he2021re}, loss re-weighting~\cite{shrivastava2016training, lin2017focal}, and transfer learning~\cite{liu2020deep, kim2020m2m}. Also in UDA, re-weighting~\cite{zou2018unsupervised, mei2020instance} and class-balanced sampling for image classification~\cite{prabhu2021sentry} were applied. We extend class-balanced sampling from classification to semantic segmentation and propose Rare Class Sampling, which addresses the co-occurrence of rare and common classes in a single semantic segmentation sample. Further, we demonstrate that re-sampling is particularly effective to train Transformers for UDA.

Li~\etal~\cite{li2017learning} have shown that knowledge distillation~\cite{hinton2015distilling} from an old task can act as a regularizer for a new task. This concept was successfully deployed with ImageNet features to semi-supervised learning~\cite{hoyer2021three} and adversarial UDA~\cite{chen2018road}. We apply this idea to self-training, show that it is particularly beneficial for Transformers, and improve it by restricting the Feature Distance to image regions with thing-classes~\cite{caesar2018coco} as ImageNet mostly labels thing-classes.

%% file: sec_methods.tex
\section{Methods}
\label{sec:methods}

\begin{figure*}
    \centering
    \includegraphics[width=\linewidth]{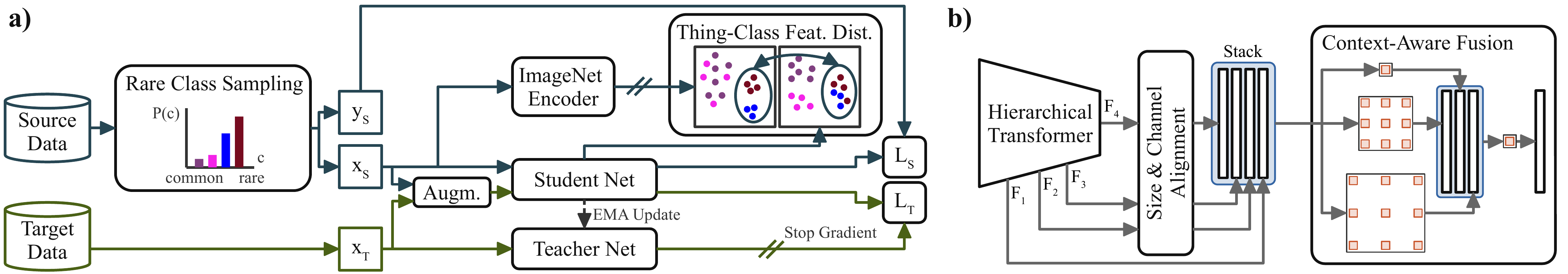}
    \caption{Overview of our UDA framework with Rare Class Sampling, Thing-Class Feature Distance, and DAFormer network.}
    \label{fig:overview}
\end{figure*}

\subsection{Self-Training (ST) for UDA}
\label{sec:methods_self_training}

First, we will give an overview over our baseline UDA method for evaluating different network architectures. 
In UDA, a neural network $g_\theta$ is trained using source domain images $\mathcal{X}_S = \{x_S^{(i)}\}_{i=1}^{N_S}$ and one-hot labels $\mathcal{Y}_S = \{y_S^{(i)}\}_{i=1}^{N_S}$ in order to achieve a good performance on target images $\mathcal{X}_T= \{x_T^{(i)}\}_{i=1}^{N_T}$ without having access to the target labels $\mathcal{Y}_T$. Naively training the network $g_\theta$ with a categorical cross-entropy (CE) loss on the source domain
\begin{equation}
    \mathcal{L}_S^{(i)} = - \sum_{j=1}^{H \times W} \sum_{c=1}^C y_S^{(i,j,c)} \log g_\theta(x_S^{(i)})^{(j,c)}
\end{equation}
usually results in a low performance on target images as the network does not generalize well to the target domain.

To address the domain gap, several strategies have been proposed that can be grouped into adversarial training~\cite{hoffman2016fcns,tsai2018learning,wang2020classes} and self-training (ST)~\cite{zou2018unsupervised,tranheden2021dacs,zhang2021prototypical} approaches. In this work, we use ST as adversarial training is known to be less stable and is currently outperformed by ST methods~\cite{tranheden2021dacs, zhang2021prototypical}.
To better transfer the knowledge from the source to the target domain, ST approaches use a teacher network $h_\phi$ (which we will describe later)
to produce pseudo-labels for the target domain data
\begin{equation}
    p_T^{(i,j,c)} = [c = \argmax_{c'} h_\phi(x_T^{(i)})^{(j,c')}]\,,
\end{equation}
where $[\cdot]$ denotes the Iverson bracket. Note that no gradients will be backpropagated into the teacher network. Additionally, a quality / confidence estimate is produced for the pseudo-labels. Here, we use the ratio of pixels exceeding a threshold $\tau$ of the maximum softmax probability~\cite{tranheden2021dacs}
\begin{equation}
    q_T^{(i)} = \frac{\sum_{j=1}^{H \times W} [\max_{c'} h_\phi(x_T^{(i)})^{(j,c')} > \tau]}{H \cdot W}\,.
\end{equation}
The pseudo-labels and their quality estimates are used to additionally train the network $g_\theta$ on the target domain
\begin{equation}
    \mathcal{L}_T^{(i)} = - \sum_{j=1}^{H \times W} \sum_{c=1}^C q_T^{(i)} p_T^{(i,j,c)} \log g_\theta(x_T^{(i)})^{(j,c)}\,.
\end{equation}

The pseudo-labels can be generated either online~\cite{araslanov2021self,tranheden2021dacs,zhou2021context} or offline~\cite{zou2018unsupervised, zou2019confidence, yang2020fda}.
We opted for online ST due to its less complex setup with only one training stage. This is important as we compare and ablate various network architectures.
In online ST, $h_\phi$ is updated based on $g_\theta$ during the training. Commonly, the weights $h_\phi$ are set as the exponentially moving average of the weights of $g_\theta$ after each training step $t$~\cite{tarvainen2017mean} to increase the stability of the predictions
\begin{equation}
    \phi_{t+1} \leftarrow \alpha \phi_t + (1 - \alpha) \theta_t\,.
\end{equation}

ST has been shown to be particularly efficient if the student network $g_\theta$ is trained on augmented target data, while the teacher network $h_\phi$ generates the pseudo-labels using non-augmented target data for semi-supervised learning~\cite{tarvainen2017mean, french2019consistency, sohn2020fixmatch} and unsupervised domain adaptation~\cite{tranheden2021dacs, araslanov2021self}
In this work, we follow DACS~\cite{tranheden2021dacs} and use color jitter, Gaussian blur, and ClassMix~\cite{olsson2020classmix} as data augmentations to learn more domain-robust features.

\subsection{DAFormer Network Architecture}
\label{sec:methods_architecture}

Previous UDA methods mostly evaluate their contributions using a (simplified) DeepLabV2 network architecture~\cite{chen2017deeplab, tsai2018learning}, which is considered to be outdated.
For that reason, we compile a network architecture that is tailored for UDA to not just achieve good supervised performance but also provide good domain-adaptation capabilities.

For the encoder, we aim for a powerful yet robust network architecture. We hypothesize that robustness is an important property in order to achieve good domain adaptation performance as it fosters the learning of domain-invariant features.
Based on recent findings~\cite{paul2021vision, bhojanapalli2021understanding, naseer2021intriguing} and an architecture comparison for UDA, which we will present in Sec.~\ref{sec:exp_comparison_networks}, Transformers~\cite{dosovitskiy2020image, touvron2021training} are a good choice for UDA as they fulfill these criteria. Although the self-attention from Transformers~\cite{vaswani2017attention} and the convolution both perform a weighted sum, their weights are computed differently: in CNNs, the weights are learned during training but fixed during testing; in the self-attention mechanism, the weights are dynamically computed based on the similarity or affinity between every pair of tokens. As a consequence, the self-similarity operation in the self-attention mechanism provides modeling means that are potentially more adaptive and general than convolution operations.

In particular, we follow the design of Mix Transformers (MiT)~\cite{xie2021segformer}, which are tailored for semantic segmentation. The image is divided into small patches of a size of $4{\times}4$  (instead of $16{\times}16$ as in ViT~\cite{dosovitskiy2020image}) in order to preserve details for semantic segmentation. To cope with the high feature resolution, sequence reduction~\cite{wang2021pyramid} is used in the self-attention blocks. The transformer encoder is designed to produce multi-level feature maps $F_i \in \mathbb{R}^{\frac{H}{2^{i+1}} \times \frac{W}{2^{i+1}} \times C_i}$. The downsampling of the feature maps is implemented by overlapping patch merging~\cite{xie2021segformer} to preserve local continuity.  

Previous works on semantic segmentation with Transformer backbones usually exploit only local information for the decoder~\cite{xie2021segformer, wang2021pyramid, zheng2021rethinking}.
In contrast, we propose to utilize additional context information in the decoder
as this has been shown to increase the robustness of semantic segmentation~\cite{kamann2021benchmarking}, a helpful property for UDA. Instead of just considering the context information of the bottleneck features~\cite{chen2017deeplab, chen2018encoder}, DAFormer uses the context across features from different encoder levels as the additional earlier features provide valuable low-level concepts for semantic segmentation at a high resolution, which can also provide important context information.
The architecture of the DAFormer decoder is shown in Fig.~\ref{fig:overview} (b). Before the feature fusion, we embed each $F_i$ to the same number of channels $C_{e}$ by a $1{\times}1$ convolution, bilinearly upsample the features to the size of $F_1$, and concatenate them. For the context-aware feature fusion, we use multiple parallel $3{\times}3$ depthwise separable convolutions~\cite{chollet2017xception} with different dilation rates~\cite{yu2015multi} and a $1{\times}1$ convolution to fuse them, similar to ASPP~\cite{chen2018encoder} but without global average pooling. In contrast to the original use of ASPP~\cite{chen2018encoder}, we do not only apply it to the bottleneck features $F_4$ but use it to fuse all stacked multi-level features. Depthwise separable convolutions have the advantage that they have a lower number of parameters than regular convolutions, which can reduce overfitting to the source domain.

\subsection{Training Strategies for UDA}
\label{sec:methods_training_strategies}

One challenge of training a more capable architecture for UDA is overfitting to the source domain. To circumvent this issue, we introduce three strategies to stabilize and regularize the UDA training: Rare Class Sampling, Thing-Class ImageNet Feature Distance, and learning rate warmup. The overall UDA framework is shown in Fig.~\ref{fig:overview} (a).

\paragraph{Rare Class Sampling (RCS)}
\label{sec:methods_rare_class_sampling}

Even though our more capable DAFormer is able to achieve better performance on difficult classes than other architectures,
we observed that the UDA performance for classes that are rare in the source dataset varies significantly over different runs.
Depending on the random seed of the data sampling order, these classes are learned at different iterations of the training or sometimes not at all as we will show in Sec.~\ref{sec:exp_rcs}.
The later a certain class is learned during the training, the worse is its performance at the end of the training. 
We hypothesize that if relevant samples containing rare classes only appear late in the training due to randomness, the network only starts to learn them later, and more importantly, it is highly possible that the network has already learned a strong bias toward common classes making it difficult to `re-learn' new concepts with very few samples. This is further reinforced as the bias is confirmed by ST with the teacher network.

To address this, we propose Rare Class Sampling (RCS). It samples images with rare classes from the source domain more often in order to learn them 
better and earlier.
The frequency $f_c$ of each class $c$ in the source dataset can be calculated based on the number of pixels with class $c$
\begin{equation}
    f_c = \frac{\sum_{i=1}^{N_S} \sum_{j=1}^{H \times W} [y_S^{(i,j,c)}]}{N_S \cdot H \cdot W}\,.
\end{equation}
The sampling probability $P(c)$ of a certain class $c$ is defined as a function of its frequency $f_c$
\begin{equation}
    P(c) = \frac{e^{(1-f_c) / T}}{\sum_{c'=1}^C e^{(1-f_{c'}) / T}}\,.
    \label{eq:P_c}
\end{equation}
Therefore, classes with a smaller frequency will have a higher sampling probability. The temperature $T$ controls the smoothness of the distribution. A higher $T$ leads to a more uniform distribution, a lower $T$ to a stronger focus on rare classes with a small $f_c$.
For each source sample, a class is sampled from the probability distribution $c \sim P$ and an image is sampled from the subset of data containing this class $x_S \sim \text{uniform}(\mathcal{X}_{S,c})$.
Eq.~\ref{eq:P_c} allows to over-sample images containing rare classes ($P(c) \geq 1 / C$ if $f_c$ is small). 
As a rare class (small $f_c$) usually co-occurs with multiple common classes (large $f_c$) in a single image, it is beneficial to sample rare classes more often than common classes ($P(c_\textit{rare}) > P(c_\textit{common})$) to get closer to a balance of the re-sampled classes.
For example, the common class road co-occurs with rare classes such as bus, train, or motorcycle and is therefore already covered when sampling images with these rare classes.
When decreasing $T$, more pixels of classes with small $f_c$ are sampled but also fewer pixels of classes with medium $f_c$. The temperature $T$ is chosen to reach a balance of the number of re-sampled pixels of classes with small and medium $f_c$ by maximizing the number of re-sampled pixels of the class with the least.

\paragraph{Thing-Class ImageNet Feature Distance (FD)}
\label{sec:methods_feature_distance}
Commonly, the semantic segmentation model $g_\theta$ is initialized with weights from ImageNet classification to start with meaningful generic features. Given that ImageNet also contains real-world images from some of the relevant high-level semantic classes, which UDA often struggles to distinguish such as train or bus, we hypothesize that the ImageNet features can provide useful guidance beyond the usual pretraining. In particular, we observe that the DAFormer network is able to segment some of the classes at the beginning of the training but forgets them after a few hundred training steps as we will show in Sec.~\ref{sec:exp_fd}.
Therefore, we assume that the useful features from ImageNet pretraining are corrupted by $\mathcal{L}_S$ and the model overfits to the synthetic source data.

In order to prevent this issue, we regularize the model based on the Feature Distance (FD) of the bottleneck features $F_\theta$ of the semantic segmentation UDA model $g_\theta$ and the bottleneck features $F_\mathit{ImageNet}$ of the ImageNet model
\begin{equation}
    d^{(i,j)} = ||F_{ImageNet}(x_S^{(i)})^{(j)} - F_\theta(x_S^{(i)})^{(j)}||_2\,.
\end{equation}

However, the ImageNet model is mostly trained on thing-classes (objects with a well-defined shape such as car or zebra) instead of stuff-classes (amorphous background regions such as road or sky)~\cite{caesar2018coco}. Therefore, we calculate the FD loss only for image regions containing thing-classes $\mathcal{C}_\mathit{things}$ described by the binary mask $M_\mathit{things}$
\begin{equation}
    \mathcal{L}_\mathit{FD}^{(i)} = \frac{\sum_{j=1}^{H_F \times W_F} d^{(i,j)} \cdot M_\mathit{things}^{(i,j)}}{\sum_{j} M_{things}^{(i,j)}}\,,
\end{equation}
This mask is obtained from the downscaled label $y_{S, \mathit{small}}$
\begin{equation}
    M_\mathit{things}^{(i,j)} = \sum_{c'=1}^C y_{S,\mathit{small}}^{i,j,c'} \cdot [c' \in \mathcal{C}_\mathit{things}]\,.
\end{equation}
To downsample the label to the bottleneck feature size, average pooling with a patch size $\frac{H}{H_F} {\times} \frac{W}{W_F}$ is applied to each class channel and a class is kept when it exceeds the ratio $r$
\begin{equation}
    y_{S,\mathit{small}}^c = [\text{AvgPool}(y_S^c, H/H_F, W/W_F) > r]\,.
\end{equation}
This ensures that only bottleneck feature pixels containing a dominant thing-class are considered for the feature distance.

The overall UDA loss $\mathcal{L}$ is the weighted sum of the presented loss components $\mathcal{L} = \mathcal{L}_S + \mathcal{L}_T + \lambda_\mathit{FD} \mathcal{L}_\mathit{FD}$.

\paragraph{Learning Rate Warmup for UDA}
\label{sec:methods_warmup}

Linearly warming up the learning rate~\cite{goyal2017accurate} at the beginning of the training has successfully been used to train both CNNs~\cite{he2016deep} and Transformers~\cite{vaswani2017attention, dosovitskiy2020image} as it improves network generalization~\cite{goyal2017accurate} by avoiding that a large adaptive learning rate variance distorts the gradient distribution at the beginning of the training~\cite{liu2019variance}. We newly introduce learning rate warmup to UDA. We posit that this is particularly important for UDA as distorting the features from ImageNet pretraining would deprive the network of useful guidance toward the real domain. During the warmup period up to iteration $t_\mathit{warm}$, the learning rate at iteration $t$ is set $\eta_t = \eta_\mathit{base} \cdot t / t_\mathit{warm}$.

%% file: sec_experiments.tex
\section{Experiments}
\label{sec:experiment}

\subsection{Implementation Details}

\noindent{\textbf{Datasets}}
For the target domain, we use the Cityscapes street scene dataset~\cite{cordts2016cityscapes} containing 2975 training and 500 validation images with resolution 2048$\times$1024.
For the source domain, we use either the GTA dataset~\cite{richter2016playing}, which contains 24,966 synthetic images with resolution 1914$\times$1052, or the Synthia dataset~\cite{ros2016synthia}, which consists of 9,400 synthetic images with resolution 1280$\times$760. As a common practice in UDA~\cite{tsai2018learning}, we resize the images to 1024$\times$512 pixels for Cityscapes and to 1280$\times$720 pixels for GTA.

\noindent{\textbf{Network Architecture}}
Our implementation is based on the mmsegmentation framework~\cite{mmseg2020}. For the DAFormer architecture, we use the MiT-B5 encoder~\cite{xie2021segformer}, which produces a feature pyramid with $C=[64, 128, 320, 512]$. The DAFormer decoder uses $C_e=256$ and dilation rates of 1, 6, 12, and 18. All encoders are pretrained on ImageNet-1k. 

\noindent{\textbf{Training}}
In accordance with~\cite{xie2021segformer, liu2021swin}, we train DAFormer with AdamW~\cite{loshchilov2018decoupled}, a learning rate of $\eta_\mathit{base} {=} 6 {\times} 10^{-5}$ for the encoder and $ 6 {\times} 10^{-4}$ for the decoder, a weight decay of $0.01$, linear learning rate warmup with $t_\mathit{warm}{=}1.5$k, and linear decay afterwards. It is trained on a batch of two $512{\times}512$ random crops for $40$k iterations.
Following DACS~\cite{tranheden2021dacs}, we use the same data augmentation parameters and set $\alpha{=}0.99$ and $\tau{=}0.968$.
The RCS temperature is set $T{=}0.01$ to maximize the sampled pixels of the class with the least pixels.
For FD, $r{=}0.75$ and $\lambda_\mathit{FD} {=} 0.005$ to induce a similar gradient magnitude into the encoder as $\mathcal{L}_S$.

\begin{table}
\centering
\caption{Comparison of the mIoU (\%) on the Cityscapes val. set of different segmentation architectures for source-only (GTA), UDA (GTA$\rightarrow$Cityscapes), and oracle (Cityscapes) training. Additionally, the relative UDA performance (Rel.) wrt. the oracle mIoU is provided. Mean and SD are calculated over 3 random seeds.}
\label{tab:basic_architecture_comparison}
\footnotesize
\begin{tabular}{lllll}
\toprule
 Architecture &       Src-Only &           UDA &         Oracle &   Rel. \\
\midrule
    DeepLabV2~\cite{chen2017deeplab} & 34.3 \spm{2.2} & 54.2 \spm{1.7} & 72.1 \spm{0.5} & 75.2\% \\
       DA Net~\cite{fu2019dual} & 30.9 \spm{2.1} & 53.7 \spm{0.2} & 72.6 \spm{0.2} & 74.0\% \\
      ISA Net~\cite{huang2019interlaced} & 32.3 \spm{2.1} & 53.3 \spm{0.4} & 72.0 \spm{0.5} & 74.0\% \\
   DeepLabV3+~\cite{chen2018encoder} & 31.0 \spm{1.4} & 53.7 \spm{1.0} & 75.6 \spm{0.9} & 71.0\% \\
    SegFormer~\cite{xie2021segformer} & \textbf{45.6} \spm{0.6} & \textbf{58.2} \spm{0.9} & \textbf{76.4} \spm{0.2} & \textbf{76.2\%} \\
\bottomrule
\end{tabular}
\end{table}

\begin{table}
\centering
\caption{Ablation of the SegFormer encoder and decoder.}
\label{tab:encoder_decoder}
\footnotesize
\begin{tabular}{lllll}
\toprule
  Encoder &             Decoder &           UDA &         Oracle &   Rel. \\
\midrule
   MiT-B5~\cite{xie2021segformer} &           SegF.~\cite{xie2021segformer} & 58.2 \spm{0.9} & 76.4 \spm{0.2} & 76.2\% \\
   MiT-B5~\cite{xie2021segformer} &           DLv3+~\cite{chen2018encoder} & 56.8 \spm{1.8} & 75.5 \spm{0.5} & 75.2\% \\
 R101~\cite{he2016deep} &           SegF.~\cite{xie2021segformer} & 50.9 \spm{1.1} & 71.3 \spm{1.3} & 71.4\% \\
 R101~\cite{he2016deep} &           DLv3+~\cite{chen2018encoder} & 53.7 \spm{1.0} & 75.6 \spm{0.9} & 71.0\% \\
\bottomrule
\end{tabular}
\end{table}

\subsection{Comparison of Network Architectures for UDA}
\label{sec:exp_comparison_networks}

First, we compare several semantic segmentation architectures with respect to their UDA performance (see Sec.~\ref{sec:methods_self_training}) on GTA$\rightarrow$Cityscapes in Tab.~\ref{tab:basic_architecture_comparison}. Additionally, we also provide the performance of the networks trained only with augmented source data (domain generalization) as well as the oracle performance trained with target labels (supervised learning).
In all cases, the model is evaluated on the Cityscapes validation set and the performance is provided as mIoU in \%. To compare how well a network is suited for UDA, we further provide the relative performance (Rel.), which normalizes the UDA mIoU by the oracle mIoU. 
Note that the oracle mIoU is generally lower than reported in the literature on supervised learning as for UDA the images of Cityscapes are downsampled by a factor of two, which is a necessary common practice in UDA to fit images from both domains and additional networks into the GPU memory.

The majority of works on UDA use DeepLabV2~\cite{chen2017deeplab} 
with ResNet-101~\cite{he2016deep} backbone. 
Interestingly, a higher oracle performance does not necessarily increase the UDA performance as can be seen for DeepLabV3+~\cite{chen2018encoder} in Tab.~\ref{tab:basic_architecture_comparison}. Generally, the studied more recent CNN architectures, do not provide a UDA performance gain over DeepLabV2. However, we identified the Transformer-based SegFormer~\cite{xie2021segformer} as a powerful architecture for UDA. It increases the mIoU for source-only / UDA / oracle training significantly 
from 34.3 / 54.2 / 72.1 to 45.6 / 58.2 / 76.4. 
We believe that especially the better domain generalization (source-only training) of SegFormer is valuable for the improved UDA performance.

To get a better insight into why SegFormer works well for UDA, we swap its encoder and decoder with ResNet101 and DeepLabV3+.
As the MiT encoder of SegFormer has an output stride of 32 but the DeepLabV3+ decoder is designed for an output stride of 8, we bilinearly upsample the SegFormer bottleneck features by $\times 4$ when combined with the DeepLabv3+ decoder.
Tab.~\ref{tab:encoder_decoder} shows that the lightweight MLP decoder of SegFormer has a slightly higher relative UDA performance (Rel.) than the heavier DLv3+ decoder (76.2\% vs 75.2\%). However, the crucial contribution to good UDA performance comes from the Transformer MiT encoder. Replacing it with the ResNet101 encoder leads to a significant performance drop of the UDA performance. Even though the oracle performance drops as well due to the smaller receptive field of the ResNet encoder~\cite{xie2021segformer}, the drop for UDA is over-proportional as shown by the relative performance decreasing from 76.2\% to 71.4\%.

\begin{table}
\centering
\caption{Influence of the encoder on UDA performance.}
\label{tab:encoder_size}
\footnotesize
\begin{tabular}{llllll}
\toprule
Enc. & Dec. &       Src-Only &           UDA &         Oracle &   Rel. \\
\midrule
    R50~\cite{he2016deep} &    DLv2~\cite{chen2017deeplab} &     29.3 & 52.1 &   70.8 & 73.6\% \\
   R101~\cite{he2016deep} &    DLv2~\cite{chen2017deeplab} &     36.9 & 53.3 &   72.5 & 73.5\% \\
    S50~\cite{zhang2020resnest} &    DLv2~\cite{chen2017deeplab} &     27.9 & 48.0 &   67.7 & 70.9\% \\
   S101~\cite{zhang2020resnest} &    DLv2~\cite{chen2017deeplab} &     35.5 & 53.5 &   72.2 & 74.1\% \\
   S200~\cite{zhang2020resnest} &    DLv2~\cite{chen2017deeplab} &     35.9 & 56.9 &   73.5 & 77.4\% \\
 MiT-B3~\cite{xie2021segformer} &   SegF.~\cite{xie2021segformer} &     42.2 & 50.8 &   76.5 & 66.4\% \\
 MiT-B4~\cite{xie2021segformer} &   SegF.~\cite{xie2021segformer} &     44.7 & 57.5 &   77.1 & 74.6\% \\
 MiT-B5~\cite{xie2021segformer} &   SegF.~\cite{xie2021segformer} &     46.2 & 58.8 &   76.2 & 77.2\% \\
\bottomrule
\end{tabular}
\end{table}

\begin{table}
\centering
\caption{Influence of learning rate warmup on UDA performance.}
\label{tab:warmup}
\footnotesize
\begin{tabular}{lllll}
\toprule
Architecture & LR Warmup &           UDA &         Oracle &   Rel. \\
\midrule
   DeepLabV2~\cite{chen2017deeplab} &      -- & 49.1 \spm{2.0} & 67.4 \spm{1.7} & 72.8\% \\
   DeepLabV2~\cite{chen2017deeplab} &     \cm & 54.2 \spm{1.7} & 72.1 \spm{0.5} & 75.2\% \\
   SegFormer~\cite{xie2021segformer} &      -- & 51.8 \spm{0.8} & 72.9 \spm{1.6} & 71.1\% \\
   SegFormer~\cite{xie2021segformer} &     \cm & 58.2 \spm{0.9} & 76.4 \spm{0.2} & 76.2\% \\
\bottomrule
\end{tabular}
\end{table}

\begin{figure}
\centering
\input{plots/tsne_imagenet_annotations}
\vspace{-0.3cm}
\resizebox{0.81\linewidth}{!}{%
\scriptsize
\setlength\tabcolsep{1pt}
{
\newcolumntype{P}[1]{>{\centering\arraybackslash}p{#1}}
\begin{tabular}{@{}*{10}{P{0.081\columnwidth}}@{}}
     {\cellcolor[rgb]{0.5,0.25,0.5}}\textcolor{white}{road} &{\cellcolor[rgb]{0.957,0.137,0.91}}sidew. &{\cellcolor[rgb]{0.275,0.275,0.275}}\textcolor{white}{build.} &{\cellcolor[rgb]{0.4,0.4,0.612}}\textcolor{white}{wall} &{\cellcolor[rgb]{0.745,0.6,0.6}}fence &{\cellcolor[rgb]{0.6,0.6,0.6}}pole &{\cellcolor[rgb]{0.98,0.667,0.118}}tr.light&{\cellcolor[rgb]{0.863,0.863,0}}sign &{\cellcolor[rgb]{0.42,0.557,0.137}}veget. & {\cellcolor[rgb]{0,0,0}}\textcolor{white}{n/a.}\\
     
     {\cellcolor[rgb]{0.596,0.984,0.596}}terrain &{\cellcolor[rgb]{0.392,0.784,0.902}}sky &{\cellcolor[rgb]{0.863,0.078,0.235}}\textcolor{white}{person} &{\cellcolor[rgb]{1,0,0}}\textcolor{white}{rider} &{\cellcolor[rgb]{0,0.353,1}}\textcolor{white}{car} &{\cellcolor[rgb]{0,0,0.275}}\textcolor{white}{truck} &{\cellcolor[rgb]{0,0.235,0.392}}\textcolor{white}{bus}&{\cellcolor[rgb]{0.549,0.902,0.078}}\textcolor{black}{train} &{\cellcolor[rgb]{0,0,0.902}}\textcolor{white}{m.bike} & {\cellcolor[rgb]{0.467,0.043,0.125}}\textcolor{white}{bike}\\
\end{tabular}
}
}
\vspace{0.2cm}
\caption{T-SNE~\cite{van2008visualizing} embedding of the bottleneck features after ImageNet pre-training for ResNet101~\cite{he2016deep} and MiT-B5~\cite{xie2021segformer} on the Cityscapes val. set, showing a better vehicle separability for MiT.}
\label{fig:tsne_imagenet}
\end{figure}
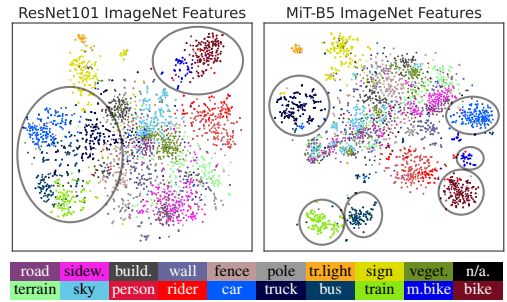

Therefore, we further investigate the influence of the encoder architecture on UDA performance. In Tab.~\ref{tab:encoder_size}, we compare different encoder designs and sizes. It can be seen that deeper models achieve a better source-only and relative performance demonstrating that deeper models generalize/adapt better to the new domain. This observation is in line with findings on the robustness of network architectures~\cite{bhojanapalli2021understanding}. Compared to CNN encoders, the MiT encoders generalize better from source-only training to the target domain. Overall, the best UDA mIoU is achieved by the MiT-B5 encoder. 
To gain insights on the improved generalization, Fig.~\ref{fig:tsne_imagenet} visualizes the ImageNet features of the target domain. Even though ResNet structures stuff-classes slightly better, MiT shines at separating semantically similar classes (e.g. all vehicle classes), which are usually particularly difficult to adapt. A possible explanation might be the texture-bias of CNNs and the shape-bias of Transformers~\cite{bhojanapalli2021understanding}.
Before we study the context-aware fusion decoder of DAFormer in Sec.~\ref{sec:exp_context_aware_fusion}, we will first discuss how to stabilize the training of MiT with the default SegFormer decoder.

\subsection{Learning Rate Warmup}

\begin{figure}
\centering
\includegraphics[width=\linewidth]{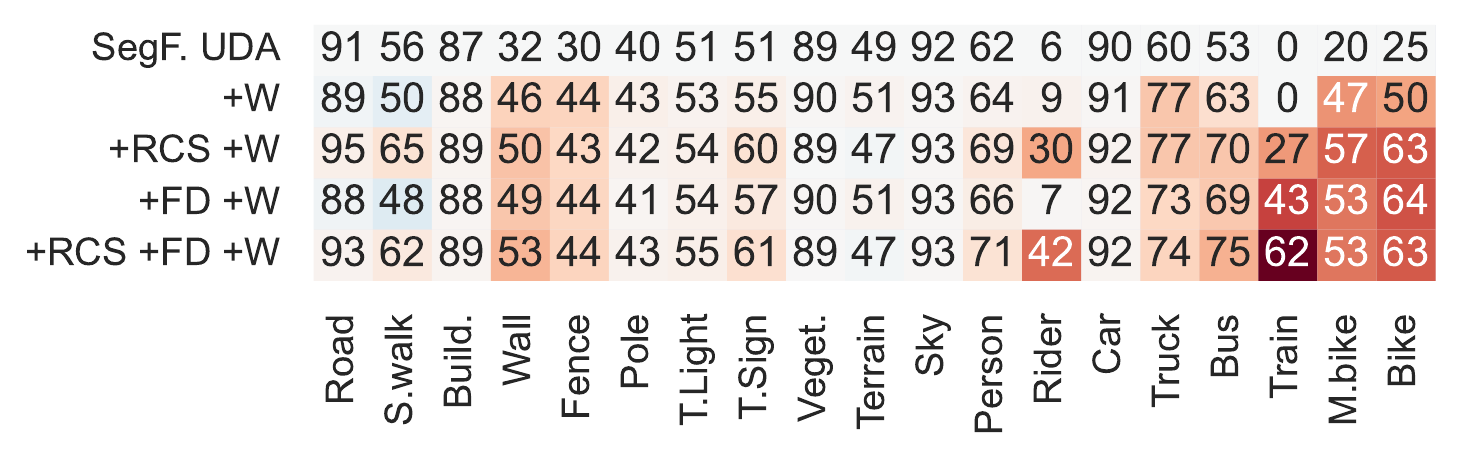}
\vspace{-0.8cm}
\caption{Comparison of the class-wise IoU of Warmup (W), RCS and FD. The color visualizes the IoU difference to the baseline.}
\label{fig:ciou_heatmap}
\end{figure}

Tab.~\ref{tab:warmup} shows that learning rate warmup significantly improves both UDA and oracle performance. 
UDA benefits even more than supervised learning from warmup (see column Rel.), showing its particular importance for UDA by stabilizing the beginning of the training, which improves difficult classes (cf. row 1 and 2 in Fig~\ref{fig:ciou_heatmap}).
As warmup is essential for a good UDA performance across different architectures, it was already applied in the previous section.

\begin{figure}
\centering
\includegraphics[width=\linewidth]{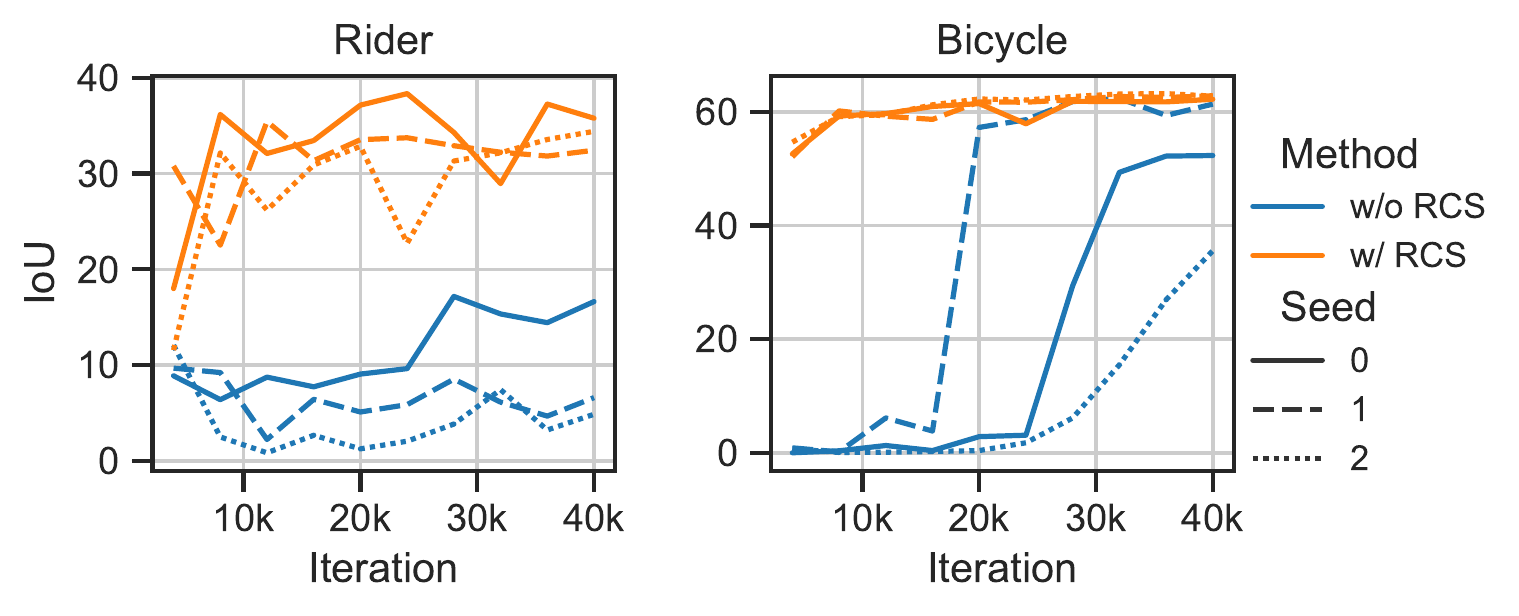}
\vspace{-0.8cm}
\caption{SegFormer UDA performance for the rare classes rider and bicycle without and with Rare Class Sampling (RCS).}
\label{fig:RCS_iou}
\end{figure}
\begin{figure}
\centering
\includegraphics[width=\linewidth]{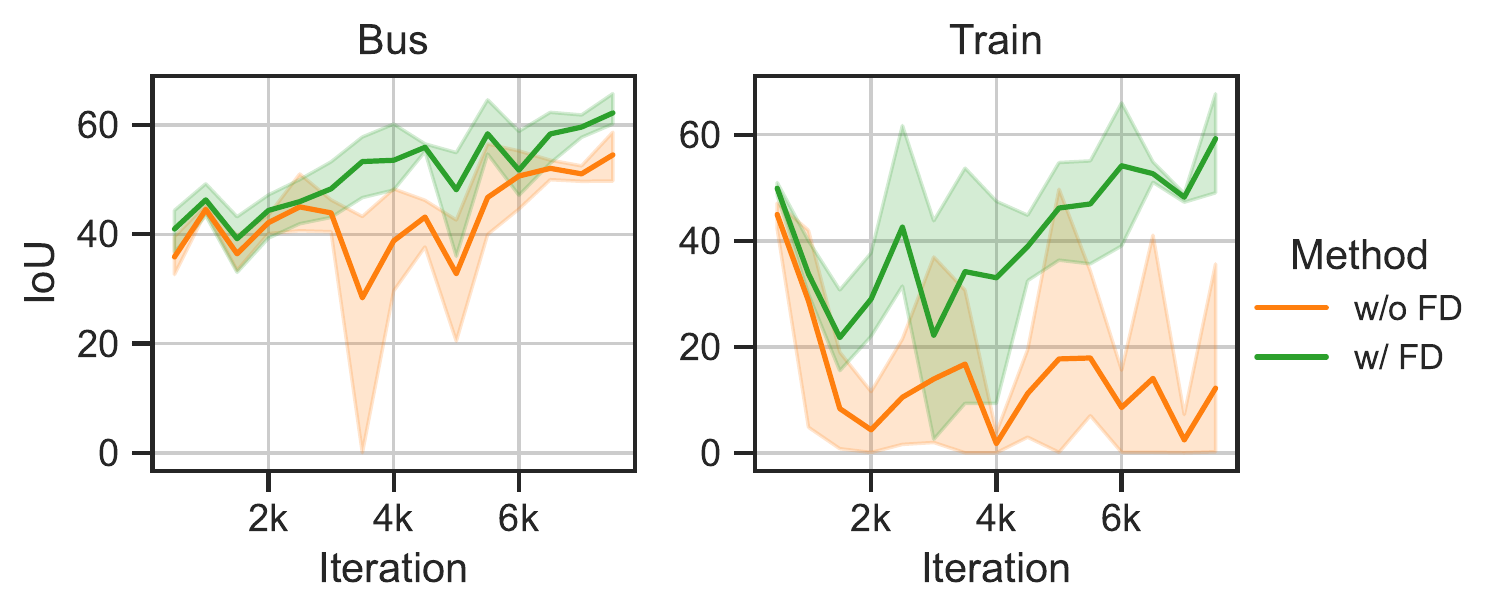}
\vspace{-0.8cm}
\caption{SegFormer UDA performance in the beginning of the training with and without ImageNet Feature Distance (FD).}
\label{fig:fd_iou}
\end{figure}

\begin{table}
\centering
\caption{Ablation of the components of the UDA framework. 
}
\label{tab:uda_improvements}
\footnotesize

\setlength{\tabcolsep}{2.5pt}
\begin{tabular}{rllllll}
\toprule
  &             Network & Warmup &           RCS &       FD & Misc. &            UDA \\
\midrule
1 & SegF.~\cite{xie2021segformer} &   -- &            -- &             -- &      -- & 51.8 \spm{0.8} \\
2 & SegF.~\cite{xie2021segformer} &  \cm &            -- &             -- &      -- & 58.2 \spm{0.9} \\
\color{gray}
3 & \color{gray}SegF.~\cite{xie2021segformer} & \color{gray}\cm & \color{gray}\cm ($T{=}\infty$) &             \color{gray}-- &      \color{gray}-- & \color{gray}62.0 \spm{1.5} \\
4 & SegF.~\cite{xie2021segformer} &  \cm &            \cm &             -- &      -- & 64.0 \spm{2.4} \\
\color{gray}5 & \color{gray}SegF.~\cite{xie2021segformer} & \color{gray}\cm &            \color{gray}-- & \color{gray}\cm (all $\mathcal{C}$) &      \color{gray}-- & \color{gray}58.8 \spm{0.4} \\
6 & SegF.~\cite{xie2021segformer} &  \cm &             -- &            \cm &      -- & 61.7 \spm{2.6} \\
7 & SegF.~\cite{xie2021segformer} &  \cm &            \cm &            \cm &      -- & 66.2 \spm{1.0} \\
8 & SegF.~\cite{xie2021segformer} &  \cm &            \cm &            \cm &     Crop PL, $\alpha {\uparrow}$ & 67.0 \spm{0.4} \\
\midrule
9 &   DLv2~\cite{chen2017deeplab} &   -- &             -- &             -- &      -- & 49.1 \spm{2.0} \\
10 &   DLv2~\cite{chen2017deeplab} &  \cm &            \cm &            \cm &     Crop PL, $\alpha {\uparrow}$ & 56.0 \spm{0.5} \\
\bottomrule
\end{tabular}

\end{table}

\begin{table*}
\centering
\caption{Comparison with state-of-the-art methods for UDA. The results for DAFormer are averaged over 3 random seeds.}
\label{tab:sota}
\setlength{\tabcolsep}{3pt}
\resizebox{\textwidth}{!}{%
\begin{tabular}{l|ccccccccccccccccccc|c}
\toprule
 & Road & S.walk & Build. & Wall & Fence & Pole & Tr.Light & Sign & Veget. & Terrain & Sky & Person & Rider & Car & Truck & Bus & Train & M.bike & Bike & mIoU\\
\toprule
\multicolumn{21}{c}{GTA5 $\to$ Cityscapes} \\
\toprule
CBST~\cite{zou2018unsupervised} & 91.8 & 53.5 & 80.5 & 32.7 & 21.0 & 34.0 & 28.9 & 20.4 & 83.9 & 34.2 & 80.9 & 53.1 & 24.0 & 82.7 & 30.3 & 35.9 & \underline{16.0} & 25.9 & 42.8 & 45.9\\
DACS~\cite{tranheden2021dacs} & 89.9 & 39.7 & \underline{87.9} & 30.7 & 39.5 & 38.5 & 46.4 & 52.8 & 88.0 & 44.0 & 88.8 & 67.2 & 35.8 & 84.5 & 45.7 & 50.2 & 0.0 & 27.3 & 34.0 & 52.1\\
CorDA\cite{wang2021domain} & \underline{94.7} & \underline{63.1} & 87.6 & 30.7 & 40.6 & 40.2 & 47.8 & 51.6 & 87.6 & \underline{47.0} & \underline{89.7} & 66.7 & 35.9 & \underline{90.2} & \underline{48.9} & 57.5 & 0.0 & 39.8 & 56.0 & 56.6\\
ProDA~\cite{zhang2021prototypical} & 87.8 & 56.0 & 79.7 & \underline{46.3} & \underline{44.8} & \underline{45.6} & \underline{53.5} & \underline{53.5} & \underline{88.6} & 45.2 & 82.1 & \underline{70.7} & \underline{39.2} & 88.8 & 45.5 & \underline{59.4} & 1.0 & \underline{48.9} & \underline{56.4} & \underline{57.5}\\
DAFormer & \textbf{95.7} & \textbf{70.2} & \textbf{89.4} & \textbf{53.5} & \textbf{48.1} & \textbf{49.6} & \textbf{55.8} & \textbf{59.4} & \textbf{89.9} & \textbf{47.9} & \textbf{92.5} & \textbf{72.2} & \textbf{44.7} & \textbf{92.3} & \textbf{74.5} & \textbf{78.2} & \textbf{65.1} & \textbf{55.9} & \textbf{61.8} & \textbf{68.3}\\

\toprule
\multicolumn{21}{c}{Synthia $\to$ Cityscapes} \\
\toprule

CBST~\cite{zou2018unsupervised} & 68.0 & 29.9 & 76.3 & 10.8 & 1.4 & 33.9 & 22.8 & 29.5 & 77.6 & -- & 78.3 & 60.6 & 28.3 & 81.6 & -- & 23.5 & -- & 18.8 & 39.8 & 42.6\\
DACS~\cite{tranheden2021dacs} & 80.6 & 25.1 & 81.9 & 21.5 & 2.9 & 37.2 & 22.7 & 24.0 & 83.7 & -- & \textbf{90.8} & 67.6 & 38.3 & 82.9 & -- & 38.9 & -- & 28.5 & 47.6 & 48.3\\
CorDA~\cite{wang2021domain} & \textbf{93.3} & \textbf{61.6} & \underline{85.3} & 19.6 & \underline{5.1} & 37.8 & 36.6 & \underline{42.8} & 84.9 & -- & \underline{90.4} & 69.7 & \underline{41.8} & 85.6 & -- & 38.4 & -- & 32.6 & \underline{53.9} & 55.0\\
ProDA~\cite{zhang2021prototypical} & \underline{87.8} & \underline{45.7} & 84.6 & \underline{37.1} & 0.6 & \underline{44.0} & \underline{54.6} & 37.0 & \textbf{88.1} & -- & 84.4 & \textbf{74.2} & 24.3 & \textbf{88.2} & -- & \underline{51.1} & -- & \underline{40.5} & 45.6 & \underline{55.5}\\
DAFormer & 84.5 & 40.7 & \textbf{88.4} & \textbf{41.5} & \textbf{6.5} & \textbf{50.0} & \textbf{55.0} & \textbf{54.6} & \underline{86.0} & -- & 89.8 & \underline{73.2} & \textbf{48.2} & \underline{87.2} & -- & \textbf{53.2} & -- & \textbf{53.9} & \textbf{61.7} & \textbf{60.9}\\
\bottomrule
\end{tabular}
}
\end{table*}

\begin{table}
\centering
\caption{Comparison of decoder architectures with MiT encoder and UDA improvements (DSC: depthwise separable convolution).}
\label{tab:decoder_fusion}
\footnotesize
\setlength{\tabcolsep}{3.5pt}
\begin{tabular}{lcrccc}
\toprule
          Decoder & $C_e$ & \#Params &          UDA &         Oracle &   Rel. \\
\midrule

        SegF.~\cite{xie2021segformer} & 768 &  3.2M & 67.0 \spm{0.4} & 76.8 \spm{0.3} & 87.2\% \\
        SegF.~\cite{xie2021segformer} & 256 &  0.5M & 67.1 \spm{1.1} & 76.5 \spm{0.4} & 87.7\% \\
       UperNet~\cite{xiao2018unified} & 512 & 29.6M & 67.4 \spm{1.1} & \textbf{78.0} \spm{0.2} & 86.4\% \\
      UperNet~\cite{xiao2018unified} & 256 &  8.3M & 66.7 \spm{1.2} & 77.4 \spm{0.3} & 86.2\% \\
ISA~\cite{huang2019interlaced} Fusion & 256 &  1.1M & 66.3 \spm{0.9} & 76.3 \spm{0.4} & 86.9\% \\
Context only at $F_4$ & 256 & 3.2M & 67.0 \spm{0.6} & 76.6 \spm{0.2} & 87.5\% \\
                     DAFormer w/o DSC & 256 & 10.0M & 67.0 \spm{1.5} & 76.7 \spm{0.6} & 87.4\% \\
                DAFormer & 256 &  3.7M & \textbf{68.3} \spm{0.5} & 77.6 \spm{0.2} & \textbf{88.0\%} \\
\bottomrule
\end{tabular}
\end{table}

\subsection{Rare Class Sampling (RCS)}
\label{sec:exp_rcs}

When training SegFormer for UDA, we observe that the performance of some classes depends on the random seed for data sampling as can be seen for the blue IoU curves in Fig.~\ref{fig:RCS_iou}. The affected classes are underrepresented in the source dataset as shown in the supplement. Interestingly, the IoU for the class bicycle starts increasing at different iterations for different seeds. We hypothesize that this is caused by the sampling order, in particular when the relevant rare classes are sampled. Further, the later the IoU starts improving, the worse is the final IoU of this class, probably due to the confirmation bias of self-training that was accumulated over earlier iterations.
Therefore, for UDA, it is especially important to learn rare classes early.

In order to address this issue, the proposed RCS increases the sampling probability of rare classes.
Fig.~\ref{fig:RCS_iou} (orange) shows that RCS results in an earlier increase of the IoU of rider/bicycle and a higher final IoU independent of the data sampling random seed. This confirms our hypothesis that an (early) sampling of rare classes is important for learning these classes properly. RCS improves the UDA performance by +5.8 mIoU (cf. row 2 and 4 in Tab.~\ref{tab:uda_improvements}).
The highest IoU increase is observed for the rare classes rider, train, motorcycle, and bicycle (cf. row 2 and 3 in Fig.~\ref{fig:ciou_heatmap}).
RCS also outperforms its special case $T = \infty$, which corresponds to `class-balanced sampling' (cf. row 3 and 4 in Tab.~\ref{tab:uda_improvements}), as class-balanced sampling does not consider the co-occurrence of multiple classes in semantic segmentation.

\subsection{Thing-Class ImageNet Feature Distance (FD)}
\label{sec:exp_fd}

While RCS gives a performance boost, the performance for thing-classes (e.g. bus and train) could still be further improved as some of the object classes that are fairly well separated in ImageNet features (see Fig.~\ref{fig:tsne_imagenet} right) are mixed together after the UDA training.
When investigating the IoU during the early training (see Fig.~\ref{fig:fd_iou} orange), we observe an early performance drop for the class train. We assume that the powerful MiT encoder overfits to the synthetic domain. When regularizing the training with the proposed FD, the performance drop is avoided (see Fig.~\ref{fig:fd_iou} green). 
Also other difficult classes such as bus, motorcycle, and bicycle benefit from the regularization (cf. row 2 and 4 in Fig.~\ref{fig:ciou_heatmap}). Overall the UDA performance is improved by +3.5 mIoU (cf. row 2 and 6 in Tab.~\ref{tab:uda_improvements}). Note that applying FD only to thing-classes, which the ImageNet features were trained on, is important for its good performance (cf. row 5 and 6).

When combining RCS and FD, we observe a further improvement to 66.2 mIoU showing that they complement each other (see row 7 in Tab.~\ref{tab:uda_improvements}). 
We notice pseudo-label drifts originating from image rectification artifacts and the ego car. As these regions are not part of the street scene segmentation task, we argue that it is not meaningful to produce pseudo-labels for them. 
Therefore, we ignore the top 15 and bottom 120 pixels of the pseudo-label. As Transformers are more expressive, we further increase $\alpha$ to 0.999 to introduce a stronger regularization from the teacher. This mitigates the pseudo-label drifts and improves the UDA mIoU
(cf. row 7 and 8 in Tab.~\ref{tab:uda_improvements}).
The overall improvement is +15.2 mIoU for SegFormer (cf. row 1 and 8) and +6.9 mIoU for DeepLabV2 (cf. row 9 and 10). When comparing both architectures, it can be seen that SegFormer benefits noticeably more, supporting our initial hypothesis that the architecture choice can limit the effectiveness of UDA methods.

\subsection{DAFormer Decoder}
\label{sec:exp_context_aware_fusion}

After regularizing and stabilizing the UDA training for a MiT encoder and a SegFormer decoder, we come back to the network architecture and investigate our DAFormer decoder with the context-aware feature fusion. Tab.~\ref{tab:decoder_fusion} shows that it improves the UDA performance over the SegFormer decoder from 67.0 to 68.3 mIoU (cf. row 1 and 8).
Further, DAFormer outperforms a variant without depthwise separable convolutions (cf. last two rows) and a variant with ISA~\cite{huang2019interlaced} instead of ASPP for feature fusion (cf. row 5 and 8). This shows that a capable but parameter-effective decoder with an inductive bias of the dilated depthwise separable convolutions is beneficial for good UDA performance.
When the context is only considered for bottleneck features, the UDA performance decreases by -1.3 mIoU (cf. row 6 and 8), revealing that the context clues from different encoder stages used in DAFormer are more domain-robust.
We further compare DAFormer to UperNet~\cite{xiao2018unified}, which iteratively upsamples and fuses the features and was used together with Transformers in~\cite{liu2021swin}. Even though UperNet achieves the best oracle performance, it is noticeably outperformed by DAFormer on UDA, which confirms that it is necessary to study and design the decoder architecture, along with the encoder architecture, specifically for UDA.

Tab.~\ref{tab:sota} shows that DAFormer outperforms previous methods by a significant margin.
On GTA$\rightarrow$Cityscapes, it improves the performance from 57.5 to 68.3 mIoU and on Synthia$\rightarrow$Cityscapes from 55.5 to 60.9 mIoU. In particular, DAFormer learns even difficult classes well, which previous methods struggled with such as train, bus, and truck.

Further details are provided in the supplement, including RCS statistics, parameter sensitivity of RCS/FD, ablation of ST, a runtime and GPU memory benchmark, a comprehensive qualitative analysis, and a discussion of limitations.

%% file: plots/tsne_imagenet_annotations.tex
\begin{tikzpicture}

	\draw (0.0, 0.0) node[inner sep=0pt] (image) {\includegraphics[width=0.8\linewidth]{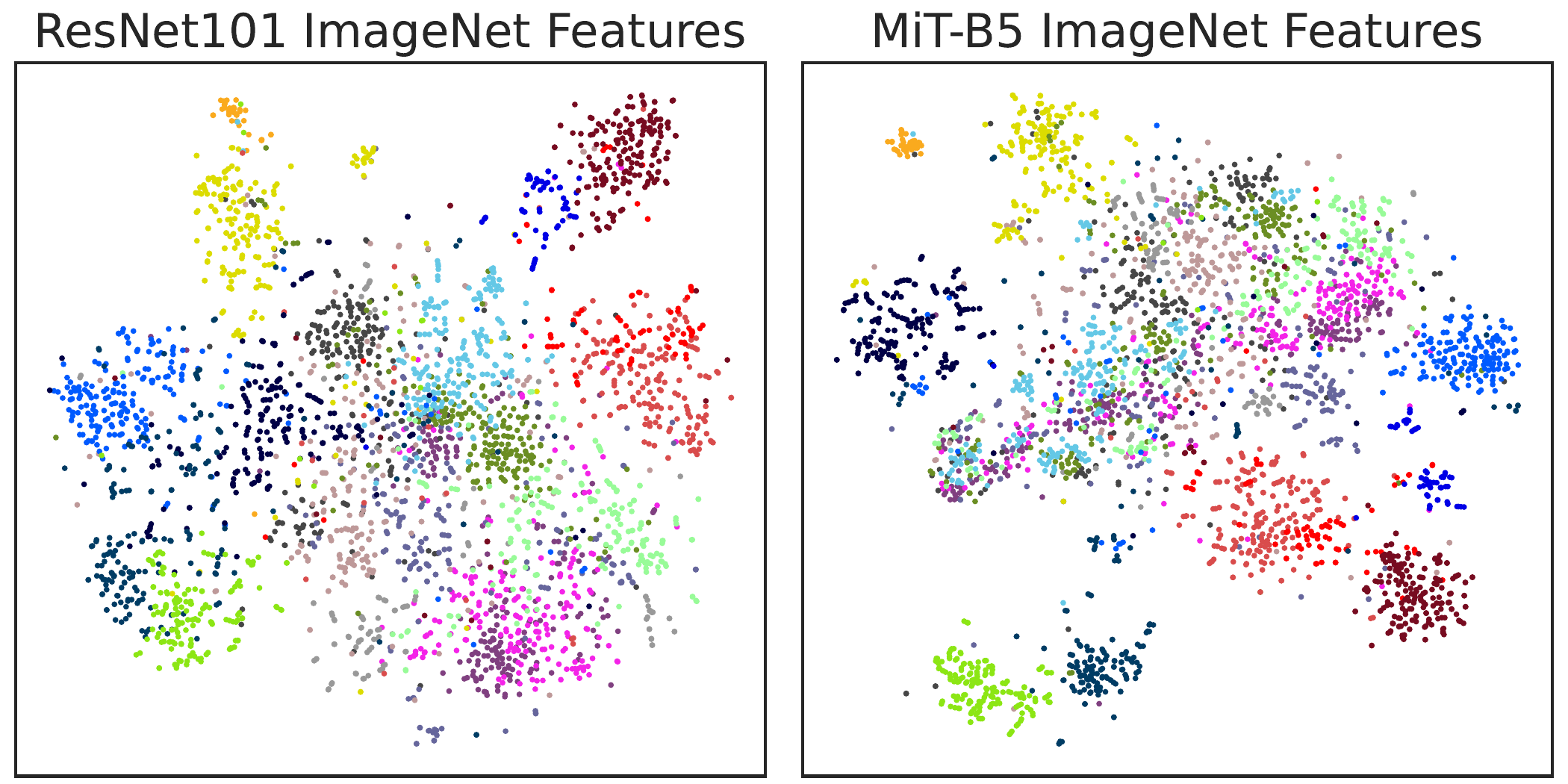}};

	\draw[black, thick, opacity=.5] (-2.5,-0.35) ellipse (0.7 and 0.9);
	
	\draw[black, thick, opacity=.5] (-0.8,0.9) ellipse (0.6 and 0.45);
	
	\draw[black, thick, opacity=.5] (0.55,0.3) ellipse (0.4 and 0.4);
	
	\draw[black, thick, opacity=.5] (0.85,-1.25) ellipse (0.3 and 0.3);
	
	\draw[black, thick, opacity=.5] (1.4,-1.15) ellipse (0.25 and 0.3);
	
	\draw[black, thick, opacity=.5] (2.85,0.17) ellipse (0.35 and 0.25);
	
	\draw[black, thick, opacity=.5] (2.7,-0.85) ellipse (0.3 and 0.3);
	
	\draw[black, thick, opacity=.5] (2.82,-0.4) ellipse (0.18 and 0.15);

\end{tikzpicture}

%% file: sec_conclusions.tex
\section{Conclusions}
\label{sec:conclusions}

We presented DAFormer, a network architecture tailored for UDA, which is based on a Transformer encoder and a context-aware fusion decoder, revealing the potential of Transformers for UDA.
Additionally, we introduced three training policies to stabilize and regularize UDA, further enabling the capabilities of DAFormer.
Overall, DAFormer represents a major advance in UDA and improves the SOTA performance by 10.8 mIoU on GTA$\rightarrow$Cityscapes and 5.4 mIoU on Synthia$\rightarrow$Cityscapes.
We would like to highlight the value of DAFormer by superseding DeepLabV2 to evaluate UDA methods on a much higher performance level.

%% file: sec_supplement.tex
\section{Overview}

The supplementary material provides further details on DAFormer as well as additional experimental results and analysis. In particular, Sec.~\ref{sec:further_implementation_details} provides further implementation details, Sec.~\ref{sec:rcs_statistics} discusses the class statistics of sampling with and without RCS, Sec.~\ref{sec:uda_dlv2} provides an ablation of the training strategies with DeepLabV2, Sec.~\ref{sec:parameter_sensitivity} studies the parameter sensitivity of RCS and FD, Sec.~\ref{sec:ablation_self_training} ablates the UDA self-training, Sec.~\ref{sec:benchmark} compares the runtime and memory consumption, Sec.~\ref{sec:qualitative_examples} analyzes example predictions, Sec.~\ref{sec:sota} compares DAFormer with additional previous UDA methods, and Sec.~\ref{sec:suppl_discussion} discusses limitations of DAFormer.

\section{Source Code and Further Details}
\label{sec:further_implementation_details}

The source code to reproduce DAFormer and all ablation studies is provided at \url{https://github.com/lhoyer/DAFormer}. Please, refer to the contained \texttt{README.md} for further information such as the environment and dataset setup.

For the distinction of thing- and stuff-classes, we follow the definition by Caesar~\etal~\cite{caesar2018coco}. Applied to Cityscapes, thing-classes are traffic light, traffic sign, person, rider, car, truck, bus, train, motorcycle, and bicycle and stuff-classes are road, sidewalk, building, wall, fence, pole, vegetation, terrain, and sky.

\section{Rare Class Sampling Statistics}
\label{sec:rcs_statistics}

\begin{figure}
\centering
\begin{subfigure}{\linewidth}
    \centering
    \includegraphics[width=\linewidth]{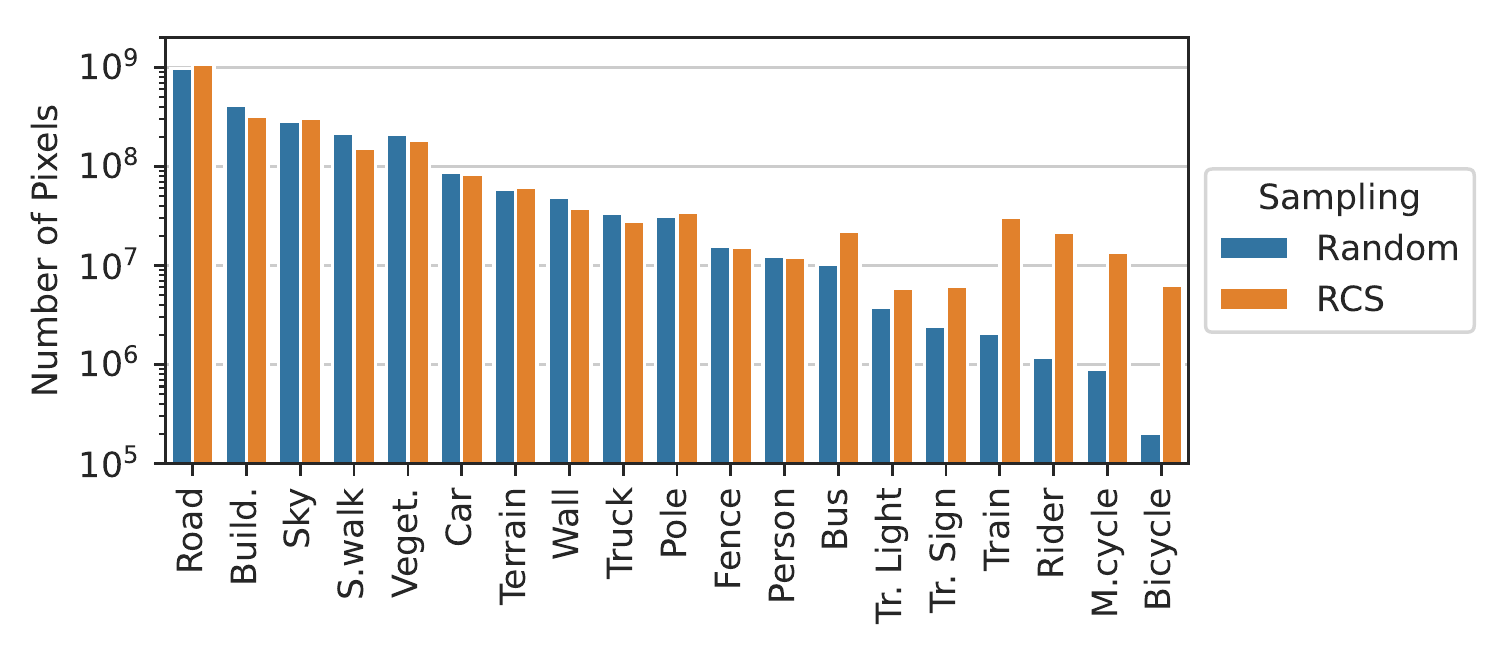}
    \caption{GTA}
    \label{fig:class_stats_gta}
\end{subfigure}
\hfill
\begin{subfigure}{\linewidth}
    \centering
    \includegraphics[width=\linewidth]{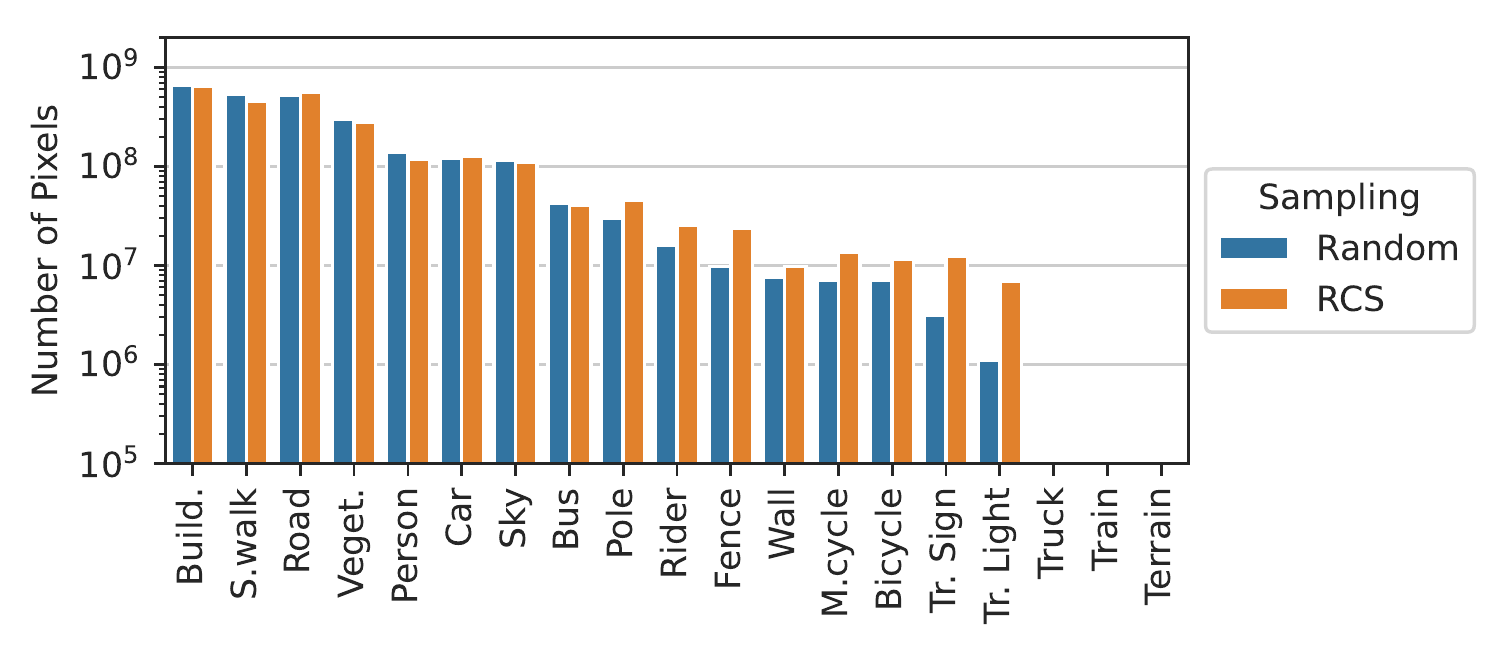}
    \caption{Synthia}
    \label{fig:class_stats_synthia}
\end{subfigure}
\caption{Class statistics of the corresponding dataset for 10k samples. Note that the y-axis is scaled \textit{logarithmically}. RCS samples images with rare classes more often than random sampling.}
\label{fig:class_stats}
\end{figure}

\begin{figure}
\centering
\begin{subfigure}{\linewidth}
    \centering
    \includegraphics[width=0.9\linewidth]{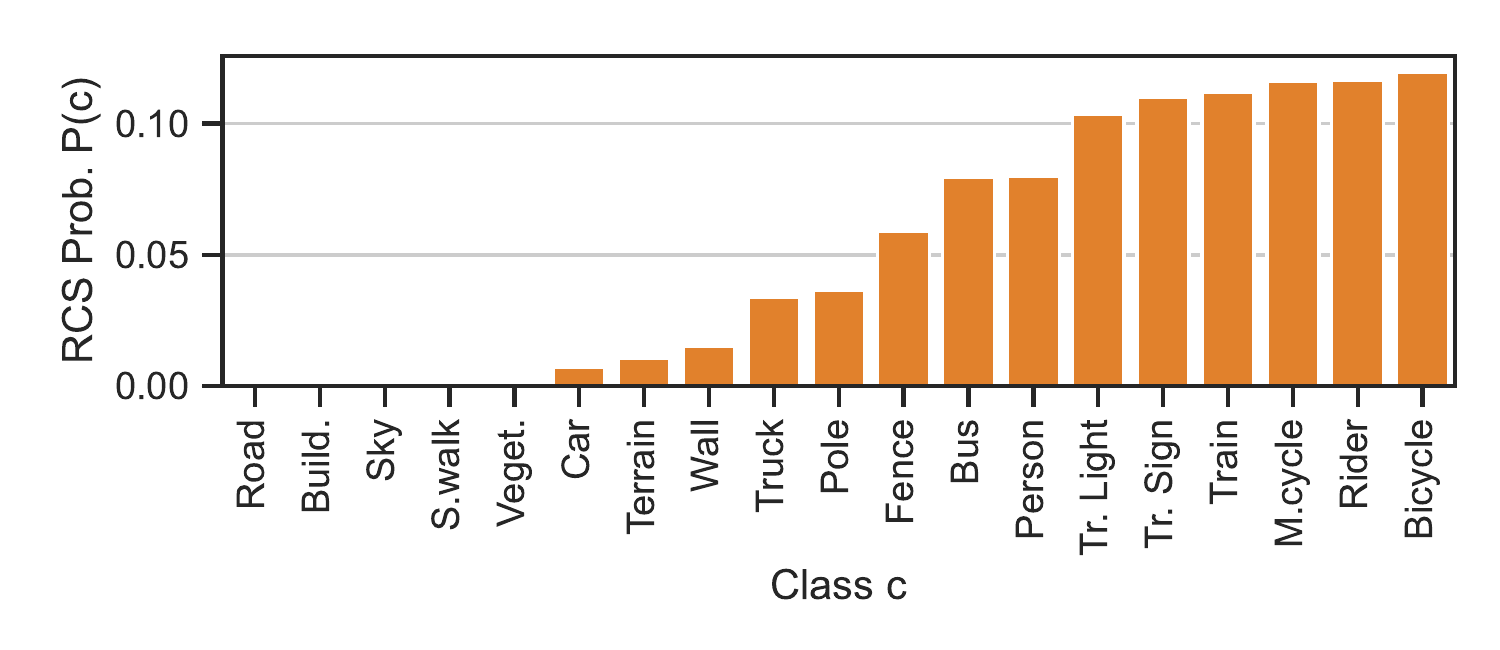}
    \caption{GTA}
    \label{fig:class_probs_gta}
\end{subfigure}
\hfill
\begin{subfigure}{\linewidth}
    \centering
    \includegraphics[width=0.9\linewidth]{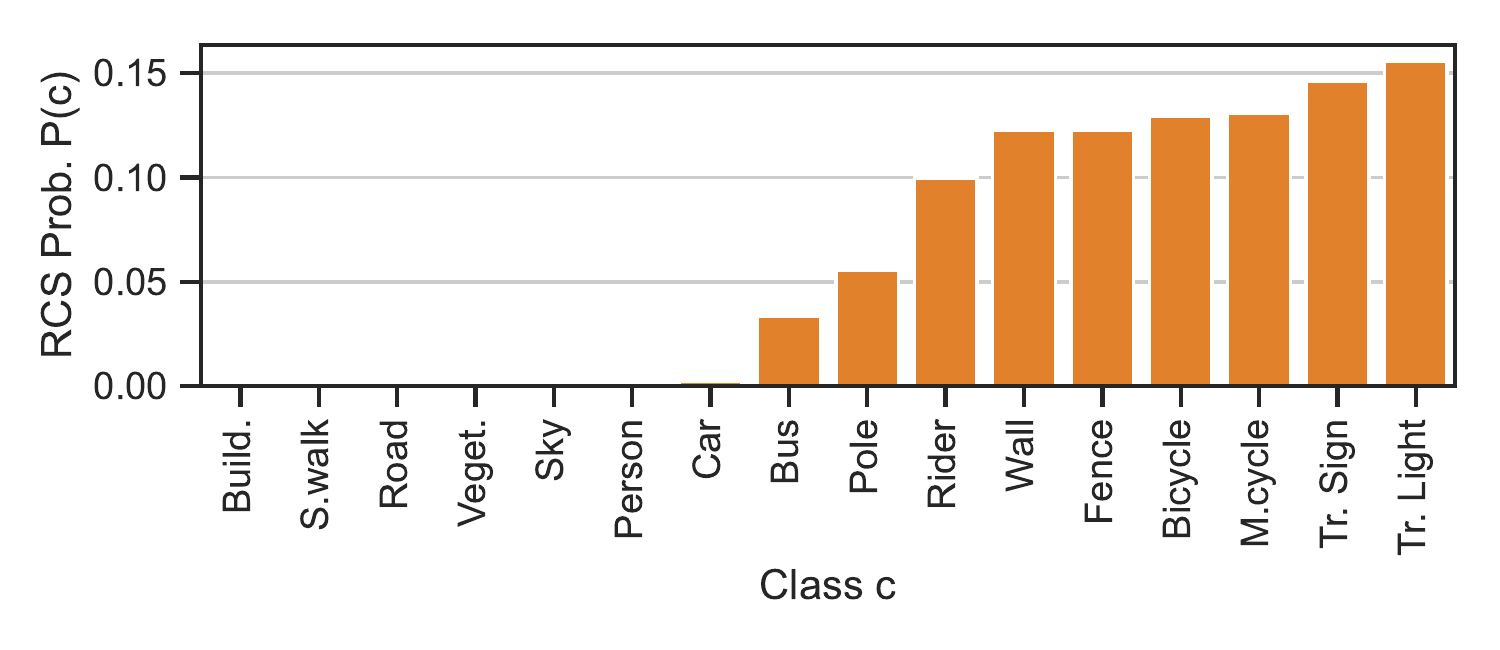}
    \caption{Synthia}
    \label{fig:class_probs_synthia}
\end{subfigure}
\caption{RCS class sampling probability $P(c)$ from Eq. 7 in the main paper with an RCS temperature of $T=0.01$.}
\label{fig:class_probs}
\end{figure}

Most (real-world) datasets have an imbalanced class distribution. This is also the case for the used source datasets as can be seen in Fig.~\ref{fig:class_stats} in the blue bars. Please note that the y-axis is scaled logarithmically so that the rare classes are still visible. This problem is addressed by RCS by sampling images with rare classes more often as discussed in Sec. 3.3 of the main paper. Therefore, more pixels of the re-sampled images belong to the rare classes as can be seen in the orange bars of Fig.~\ref{fig:class_stats}, which directly results in a significantly improved IoU for these rare classes as can be seen in Fig.~6 of the main paper. The RCS temperature $T=0.01$ is chosen to (approximately) maximize the number of re-sampled pixels of the class with the least re-sampled pixels as described in Sec. 3.3 of the main paper. This strategy results in a balance of the number of re-sampled pixels of the classes with the least re-sampled pixels as can be seen in the orange bars of Fig.~\ref{fig:class_stats_gta} for the classes \emph{traffic light} and \emph{bicycle}.

Further, Fig.~\ref{fig:class_probs} shows the RCS class sampling probabilities $P(c)$ from Eq. 7 in the main paper for the default RCS temperature $T=0.01$. It can be seen that the class sampling probability for rare classes is higher than for common classes as expected. For some common classes such as \emph{road} and \emph{sky}, $P(c)$ is very close to zero. As these common classes are part of almost every image, it is not necessary to specifically sample images containing these particular common classes.

\section{Training Strategies for DeepLabV2}
\label{sec:uda_dlv2}

\begin{table}
\centering
\caption{Ablation of the components of the UDA framework for SegFormer and DeepLabV2. Mean and standard deviation are calculated over 3 random seeds.}
\label{tab:uda_dlv2}
\footnotesize
\begin{tabular}{ccccc}
\toprule
    Network & Warmup & RCS &  FD &   mIoU \\
\midrule
SegF.~\cite{xie2021segformer} &     -- &  -- &  -- & 51.8 \spm{0.8} \\
SegF.~\cite{xie2021segformer} &    \cm &  -- &  -- & 58.2 \spm{0.9} \\
SegF.~\cite{xie2021segformer} &    \cm &  -- & \cm & 61.7 \spm{2.6} \\
SegF.~\cite{xie2021segformer} &    \cm & \cm &  -- & 64.0 \spm{2.4} \\
SegF.~\cite{xie2021segformer} &    \cm & \cm & \cm & 66.2 \spm{1.0} \\
 \hline
  DLv2~\cite{chen2017deeplab} &     -- &  -- &  -- & 49.1 \spm{2.0} \\
  DLv2~\cite{chen2017deeplab} &    \cm &  -- &  -- & 54.2 \spm{1.7} \\
  DLv2~\cite{chen2017deeplab} &    \cm &  -- & \cm & 55.5 \spm{1.3} \\
  DLv2~\cite{chen2017deeplab} &    \cm & \cm &  -- & 55.7 \spm{0.5} \\
  DLv2~\cite{chen2017deeplab} &    \cm & \cm & \cm & 56.6 \spm{1.2} \\
\bottomrule
\end{tabular}
\end{table}

Tab.~\ref{tab:uda_dlv2} shows the performance of the improved UDA training strategies for the DeepLabV2 architecture in addition to the SegFormer architecture from the main paper. It demonstrates that learning rate warmup, RCS, FD, and their combination are all beneficial for DeepLabV2 as well. However, the performance improvement for SegFormer is significantly larger than for DeepLabV2, supporting our hypothesis that the network architecture is crucial for UDA performance.

\section{Parameter Sensitivity of RCS and FD}
\label{sec:parameter_sensitivity}

\begin{table}
\centering
\caption{Hyperparameter sensitivity study for RCS and FD. The default parameters are marked with *. The evaluation setup is equivalent to row 8 in Tab.~5 of the main paper.}
\label{tab:parameter_sensitivity}
\footnotesize
\begin{tabular}{lc}
\toprule
RCS $T$ & mIoU \\
\midrule
  0.001 &                          65.6 \\
  0.002 &                          66.8 \\
  0.01* &                          66.8 \\
   0.05 &                          66.8 \\
    0.1 &                          65.5 \\
\bottomrule
\end{tabular}
\quad
\begin{tabular}{lc}
\toprule
$\lambda_\mathit{FD}$ & mIoU \\
\midrule
         0.001 &                          65.8 \\
         0.002 &                          66.2 \\
        0.005* &                          66.8 \\
          0.01 &                          66.5 \\
          0.02 &                          65.8 \\
\bottomrule
\end{tabular}
\end{table}

To analyze the sensitivity of the parameters of FD and RCS, Tab.~\ref{tab:parameter_sensitivity} shows a study of $T$ for RCS and $\lambda_\mathit{FD}$ for FD. It can be seen that RCS is stable up to a deviation of factor 5 from the default value. For FD, the weighting is stable up to a factor of about 2. Given that both default values are chosen according to an intuitive strategy (maximization of re-sampled pixels for the class with the least re-sampled pixels for $T$ and gradient magnitude balance for $\lambda_\mathit{FD}$), the robust range of the hyperparameters is sufficient to select a good value according to the described strategy.

\section{Ablation of Self-Training}
\label{sec:ablation_self_training}

\begin{table}
\centering
\caption{Ablation of UDA self-training (ST, see Sec.~3.1 in the main paper) with and without RCS and FD. Warmup is enabled in all configuration. The experiments are conducted with SegFormer~\cite{xie2021segformer} and complement Tab. 5 in the main paper.}
\label{tab:ablation_self_training}
\footnotesize
\begin{tabular}{lcc}
\toprule
 & w/o (RCS+FD) & w/ (RCS+FD) \\
\midrule
w/o ST & 45.6\spm{0.6} & 50.7\spm{0.3} \\
w/ ST  & 58.2\spm{0.9} & 66.2\spm{1.0} \\
\bottomrule
\end{tabular}
\end{table}

As FD and RCS operate on source data, they can also be used to improve the domain generalization ability of a model trained only on the source domain without self-training (ST) on the target domain. Without ST (row 1 in Tab.~\ref{tab:ablation_self_training}), RCS and FD increase the network performance by +5.1 mIoU, demonstrating their benefit for domain generalization. Combined with ST (row 2 in Tab.~\ref{tab:ablation_self_training}), their improvement even increases to +8.0 mIoU showing that RCS and FD reinforce ST, confirming their particular importance for UDA as well.

\section{Runtime and Memory Consumption}
\label{sec:benchmark}

\begin{table}
\centering
\footnotesize
\caption{Runtime and memory consumption of different network architectures and UDA methods on a single RTX 2080 Ti GPU.}
\label{tab:benchmark}
\setlength\tabcolsep{2.5pt}
\begin{tabular}{llrrrr}
\toprule
UDA Method & Network   & \begin{tabular}[c]{@{}r@{}}Training \\ Throughput \\(it/s)\end{tabular} & \begin{tabular}[c]{@{}r@{}}Inference\\ Throughput \\(img/s)\end{tabular} & \begin{tabular}[c]{@{}r@{}}Training\\ GPU\\ Memory\end{tabular} & \begin{tabular}[c]{@{}r@{}}Num.\\ Params\end{tabular} \\
\midrule
ST         & DLv2~\cite{chen2017deeplab}   & 1.24 & 11.3 & 5.6 GB & 43.2M \\
ST         & SegF.~\cite{xie2021segformer} & 0.95 & 8.9  & 7.7 GB & 84.6M \\
ST+RCS+FD  & DLv2~\cite{chen2017deeplab}   & 0.91 & 11.3 & 8.6 GB & 43.2M \\
ST+RCS+FD  & SegF.~\cite{xie2021segformer} & 0.75 & 8.9  & 8.8 GB & 84.6M \\
ST+RCS+FD  & DAFormer                      & 0.71 & 8.7  & 9.6 GB & 85.2M \\
\bottomrule
\end{tabular}
\end{table}

DAFormer can be trained on a single RTX 2080 Ti GPU within 16 hours (0.7 it/s) while requiring about 9.6 GB GPU memory during training. It has a throughput of 8.7 img/s for inference. In Tab.~\ref{tab:benchmark}, DAFormer is compared with other network architectures and UDA methods with respect to runtime and memory consumption. Even though DAFormer is heavier than DeepLabV2 and SegFormer, it requires only 12\% more GPU memory and about 30\% more training/inference time than DeepLabV2 when the same UDA configuration is used. When ablating the proposed RCS and FD, the GPU memory consumption is further reduced by 54\% and the training time is decreased by 36\% mainly due to the additional ImageNet encoder and the feature distance calculation. However, as this is only relevant for training, the inference throughput is the same.

\section{Qualitative Analysis}
\label{sec:qualitative_examples}

\begin{figure}
\centering
\begin{subfigure}{\linewidth}
    \includegraphics[width=\linewidth]{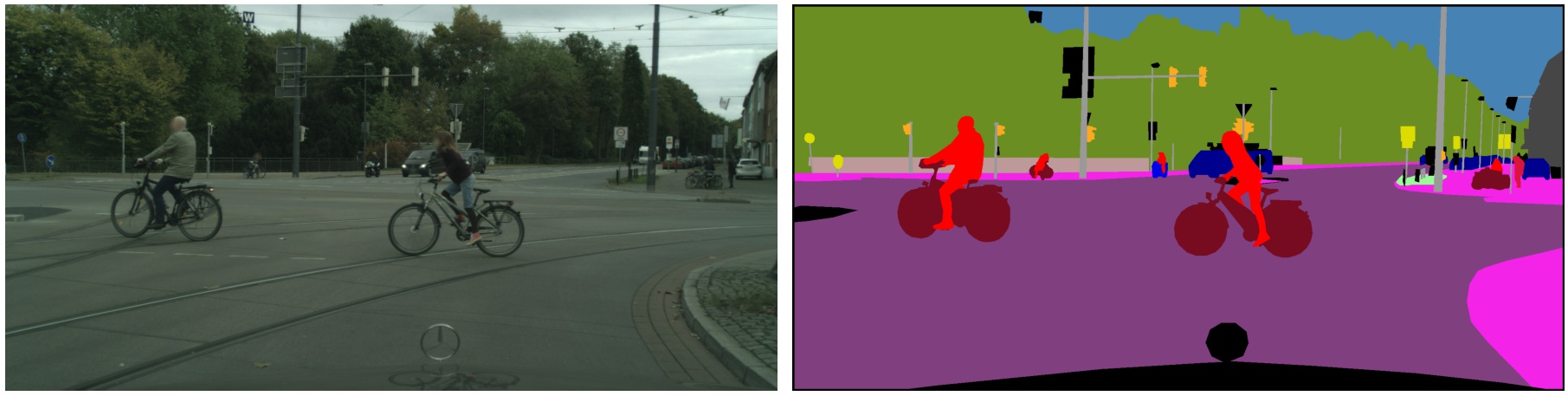}
    \caption{Bicycle annotation policy on Cityscapes}
    \label{fig:label_policy_bicycle_cityscapes}
\end{subfigure}
\hfill
\begin{subfigure}{\linewidth}
    \includegraphics[width=\linewidth]{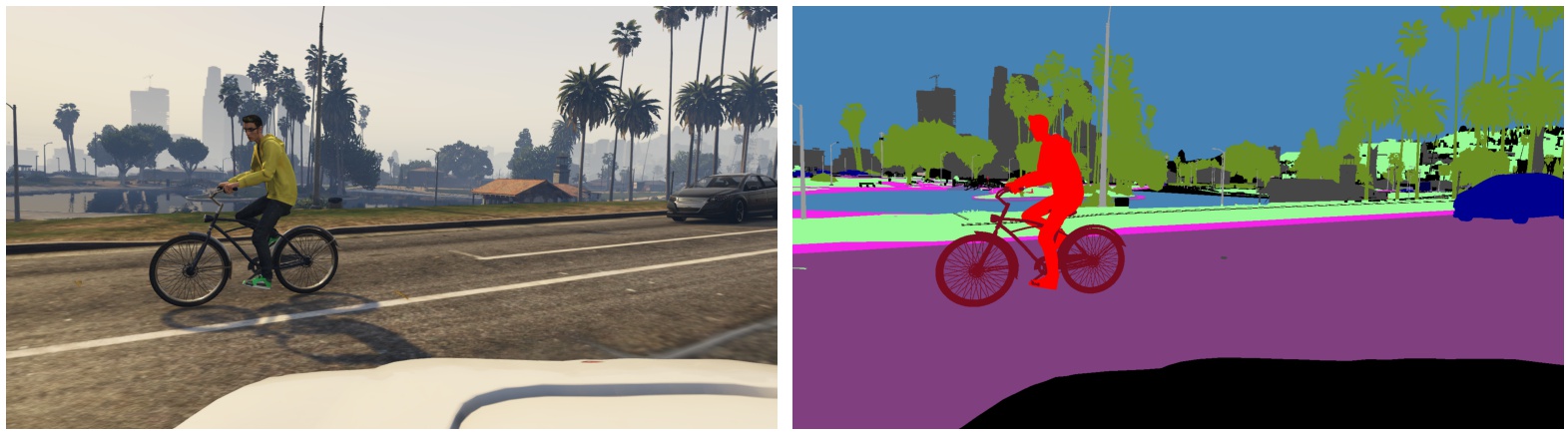}
    \caption{Bicycle annotation policy on GTA}
    \label{fig:label_policy_bicycle_gta}
\end{subfigure}
\caption{Different annotation policies for bicycle on Cityscapes and GTA. It can be seen that the entire wheel is segmented as bicycle for Cityscapes, while only the tire and spokes of the wheel are segmented as bicycle for GTA.}
\label{fig:label_policy_bicycle}
\end{figure}

\begin{figure*}
\centering
\input{preds/prediction_head}
\input{preds/8}
\input{preds/91}
\input{preds/116}
\input{preds/228} 
\input{preds/palette}
\vspace{-0.15cm}
\caption{Example predictions showing a better recognition of \emph{train} as opposed to \emph{bus} by DAFormer on GTA$\rightarrow$Cityscapes.}
\label{fig:predictions_train}
\vspace{\floatsep}

\centering
\input{preds/prediction_head}
\input{preds/51}
\input{preds/139}
\input{preds/200}
\input{preds/454} 
\vspace{-0.5cm}
\caption{Example predictions showing a better recognition of \emph{bus} by DAFormer on GTA$\rightarrow$Cityscapes.}
\label{fig:predictions_bus}
\vspace{\floatsep}

\centering
\input{preds/prediction_head}
\input{preds/21}
\input{preds/134} 
\input{preds/268}
\input{preds/429}
\vspace{-0.5cm}
\caption{Example predictions showing a better recognition of \emph{truck} by DAFormer on GTA$\rightarrow$Cityscapes.}
\label{fig:predictions_truck}
\end{figure*}

\begin{figure*}
\centering
\input{preds/prediction_head}
\input{preds/76}
\input{preds/133}
\input{preds/155}
\input{preds/364} 
\input{preds/palette}
\vspace{-0.15cm}
\caption{Example predictions showing a better recognition of \emph{car} as opposed to \emph{truck} by DAFormer on GTA$\rightarrow$Cityscapes.}
\label{fig:predictions_car}
\vspace{\floatsep}

\centering
\input{preds/prediction_head}
\input{preds/23}
\input{preds/80}
\input{preds/151}
\input{preds/349} 
\vspace{-0.5cm}
\caption{Example predictions showing a better recognition of \emph{sidewalk} as opposed to \emph{road} by DAFormer on GTA$\rightarrow$Cityscapes.}
\label{fig:predictions_sidewalk}
\vspace{\floatsep}

\centering
\input{preds/prediction_head}
\input{preds/424}
\input{preds/90}
\input{preds/366}
\input{preds/455}
\vspace{-0.5cm}
\caption{Further example predictions on GTA$\rightarrow$Cityscapes showing a better recognition of \emph{bicycle}, \emph{rider}, and \emph{fence}.}
\label{fig:predictions_other}
\end{figure*}

\begin{figure*}
\centering
\input{preds/prediction_head}
\input{preds/19}
\input{preds/100}
\input{preds/palette}
\vspace{-0.15cm}
\caption{Typical error cases on GTA$\rightarrow$Cityscapes: Confusion of \emph{sidewalk} and \emph{road} if the texture is similar.}
\label{fig:error_cases_sidewalk}
\vspace{\floatsep}

\centering
\input{preds/prediction_head}
\input{preds/278}
\input{preds/257}
\vspace{-0.5cm}
\caption{Typical error cases on GTA$\rightarrow$Cityscapes: Confusion of \emph{wall} and \emph{fence}.}
\label{fig:error_cases_wall}
\vspace{\floatsep}

\centering
\input{preds/prediction_head}
\input{preds/438}
\input{preds/201}
\vspace{-0.5cm}
\caption{Typical error cases on GTA$\rightarrow$Cityscapes: Misclassification of special vans.}
\label{fig:error_cases_vans}
\vspace{\floatsep}

\centering
\input{preds/prediction_head}
\input{preds/71}
\input{preds/357}
\vspace{-0.5cm}
\caption{Typical error cases on GTA$\rightarrow$Cityscapes: Misclassification of  partly-occluded \emph{busses}.}
\label{fig:error_cases_occluded_bus}
\vspace{\floatsep}

\centering
\input{preds/prediction_head}
\input{preds/29}
\input{preds/38}
\vspace{-0.5cm}
\caption{Typical error cases on GTA$\rightarrow$Cityscapes: Misclassification of \emph{persons} close to bikes (top row), standing \emph{riders} (bottom row), and missing segmentation of the inside of the \emph{bicycle} wheel (bottom row).}
\label{fig:error_cases_rider}
\end{figure*}

\paragraph{Comparison with ProDA} The better performance of DAFormer compared to ProDA~\cite{zhang2021prototypical} is also reflected in example predictions shown in column 4 and 5 of Fig.~\ref{fig:predictions_train}-\ref{fig:predictions_other}. The major improvements come from a better recognition of the classes \emph{train} (Fig.~\ref{fig:predictions_train}), \emph{bus} (Fig.~\ref{fig:predictions_bus}), \emph{truck} (Fig.~\ref{fig:predictions_truck}), \emph{car} (Fig.~\ref{fig:predictions_car}), and \emph{sidewalk} (Fig.~\ref{fig:predictions_sidewalk}) across different perspectives, object sizes, and appearances. Fig.~\ref{fig:predictions_other} also shows a better recognition of \emph{rider}, \emph{bicycle}, and \emph{fence}.
Furthermore, in the shown examples, it can generally be seen that DAFormer better segments fine structures, which is especially beneficial for small classes such as \emph{pole}, \emph{traffic sign}, and \emph{traffic light}.

\paragraph{Domain Generalization} Additionally, column 2 and 3 of Fig.~\ref{fig:predictions_train}-\ref{fig:predictions_other} compare the source-only training of DeepLabV2 and SegFormer. It can be seen that SegFormer better generalizes from the source training to the target domain than DeepLabV2. Still, there is a considerable gap between SegFormer with source-only training and DAFormer, showing that the adaptation to the target domain is essential.

\paragraph{Error Cases} To give additional insights into the limitations of DAFormer, Fig.~\ref{fig:error_cases_sidewalk}-\ref{fig:error_cases_rider} show some of the typical error cases. This includes confusion of \emph{sidewalk} and \emph{road} if the texture is similar or there are cycle path markings (Fig.~\ref{fig:error_cases_sidewalk}), confusion of \emph{wall} and \emph{fence} (Fig.~\ref{fig:error_cases_wall}), misclassification of some special vans (Fig.~\ref{fig:error_cases_vans}), misclassification of partly-occluded \emph{busses} (Fig.~\ref{fig:error_cases_occluded_bus}), misclassification of \emph{persons} close to bikes (Fig.~\ref{fig:error_cases_rider} top), misclassification of standing \emph{riders} (Fig.~\ref{fig:error_cases_rider} bottom), and segmentation of only the bicycle tires (Fig.~\ref{fig:error_cases_rider} bottom).
That only the tires and not the inside of the wheel of a bicycle are segmented is caused by the annotation policy of GTA (see Fig.~\ref{fig:label_policy_bicycle}). Note that for many of the error cases, also previous methods such as ProDA experience similar issues.

\section{Comparison with Previous Methods}
\label{sec:sota}

\begin{table*}
\vspace{1.4cm}
\centering
\caption{Comparison with previous methods on GTA$\rightarrow$Cityscapes. Our results (DAFormer) are averaged over 3 random seeds.}
\label{tab:sota_gta}
\setlength{\tabcolsep}{3pt}
\resizebox{\textwidth}{!}{%
\begin{tabular}{l|ccccccccccccccccccc|c}
\toprule
 & Road & S.walk & Build. & Wall & Fence & Pole & Tr.Light & Sign & Veget. & Terrain & Sky & Person & Rider & Car & Truck & Bus & Train & M.bike & Bike & mIoU\\
\midrule
AdaptSeg~\cite{tsai2018learning} & 86.5 & 25.9 & 79.8 & 22.1 & 20.0 & 23.6 & 33.1 & 21.8 & 81.8 & 25.9 & 75.9 & 57.3 & 26.2 & 76.3 & 29.8 & 32.1 & 7.2 & 29.5 & 32.5 & 41.4\\
CyCADA~\cite{hoffman2018cycada} & 86.7 & 35.6 & 80.1 & 19.8 & 17.5 & 38.0 & 39.9 & 41.5 & 82.7 & 27.9 & 73.6 & 64.9 & 19.0 & 65.0 & 12.0 & 28.6 & 4.5 & 31.1 & 42.0 & 42.7\\
CLAN~\cite{luo2019taking} & 87.0 & 27.1 & 79.6 & 27.3 & 23.3 & 28.3 & 35.5 & 24.2 & 83.6 & 27.4 & 74.2 & 58.6 & 28.0 & 76.2 & 33.1 & 36.7 & 6.7 & 31.9 & 31.4 & 43.2\\
ADVENT~\cite{vu2019advent} & 89.4 & 33.1 & 81.0 & 26.6 & 26.8 & 27.2 & 33.5 & 24.7 & 83.9 & 36.7 & 78.8 & 58.7 & 30.5 & 84.8 & 38.5 & 44.5 & 1.7 & 31.6 & 32.4 & 45.5\\
APODA~\cite{yang2020adversarial} & 85.6 & 32.8 & 79.0 & 29.5 & 25.5 & 26.8 & 34.6 & 19.9 & 83.7 & 40.6 & 77.9 & 59.2 & 28.3 & 84.6 & 34.6 & 49.2 & 8.0 & 32.6 & 39.6 & 45.9\\
CBST~\cite{zou2018unsupervised} & 91.8 & 53.5 & 80.5 & 32.7 & 21.0 & 34.0 & 28.9 & 20.4 & 83.9 & 34.2 & 80.9 & 53.1 & 24.0 & 82.7 & 30.3 & 35.9 & 16.0 & 25.9 & 42.8 & 45.9\\
PatchAlign~\cite{tsai2019domain} & 92.3 & 51.9 & 82.1 & 29.2 & 25.1 & 24.5 & 33.8 & 33.0 & 82.4 & 32.8 & 82.2 & 58.6 & 27.2 & 84.3 & 33.4 & 46.3 & 2.2 & 29.5 & 32.3 & 46.5\\
MRKLD~\cite{zou2019confidence} & 91.0 & 55.4 & 80.0 & 33.7 & 21.4 & 37.3 & 32.9 & 24.5 & 85.0 & 34.1 & 80.8 & 57.7 & 24.6 & 84.1 & 27.8 & 30.1 & 26.9 & 26.0 & 42.3 & 47.1\\
BDL~\cite{li2019bidirectional} & 91.0 & 44.7 & 84.2 & 34.6 & 27.6 & 30.2 & 36.0 & 36.0 & 85.0 & 43.6 & 83.0 & 58.6 & 31.6 & 83.3 & 35.3 & 49.7 & 3.3 & 28.8 & 35.6 & 48.5\\
FADA~\cite{wang2020classes} & 91.0 & 50.6 & 86.0 & 43.4 & 29.8 & 36.8 & 43.4 & 25.0 & 86.8 & 38.3 & 87.4 & 64.0 & 38.0 & 85.2 & 31.6 & 46.1 & 6.5 & 25.4 & 37.1 & 50.1\\
CAG~\cite{zhang2019category} & 90.4 & 51.6 & 83.8 & 34.2 & 27.8 & 38.4 & 25.3 & 48.4 & 85.4 & 38.2 & 78.1 & 58.6 & 34.6 & 84.7 & 21.9 & 42.7 & 41.1 & 29.3 & 37.2 & 50.2\\
Seg-Uncert.~\cite{zheng2021rectifying} & 90.4 & 31.2 & 85.1 & 36.9 & 25.6 & 37.5 & 48.8 & 48.5 & 85.3 & 34.8 & 81.1 & 64.4 & 36.8 & 86.3 & 34.9 & 52.2 & 1.7 & 29.0 & 44.6 & 50.3\\
FDA\cite{yang2020fda} & 92.5 & 53.3 & 82.4 & 26.5 & 27.6 & 36.4 & 40.6 & 38.9 & 82.3 & 39.8 & 78.0 & 62.6 & 34.4 & 84.9 & 34.1 & 53.1 & 16.9 & 27.7 & 46.4 & 50.5\\
PIT\cite{lv2020cross} & 87.5 & 43.4 & 78.8 & 31.2 & 30.2 & 36.3 & 39.9 & 42.0 & 79.2 & 37.1 & 79.3 & 65.4 & 37.5 & 83.2 & 46.0 & 45.6 & 25.7 & 23.5 & 49.9 & 50.6\\
IAST\cite{mei2020instance} & 93.8 & 57.8 & 85.1 & 39.5 & 26.7 & 26.2 & 43.1 & 34.7 & 84.9 & 32.9 & 88.0 & 62.6 & 29.0 & 87.3 & 39.2 & 49.6 & 23.2 & 34.7 & 39.6 & 51.5\\
DACS~\cite{tranheden2021dacs} & 89.9 & 39.7 & \underline{87.9} & 30.7 & 39.5 & 38.5 & 46.4 & 52.8 & 88.0 & 44.0 & 88.8 & 67.2 & 35.8 & 84.5 & 45.7 & 50.2 & 0.0 & 27.3 & 34.0 & 52.1\\
SAC~\cite{araslanov2021self} & 90.4 & 53.9 & 86.6 & 42.4 & 27.3 & 45.1 & 48.5 & 42.7 & 87.4 & 40.1 & 86.1 & 67.5 & 29.7 & 88.5 & \underline{49.1} & 54.6 & 9.8 & 26.6 & 45.3 & 53.8\\
CTF~\cite{ma2021coarse} & 92.5 & 58.3 & 86.5 & 27.4 & 28.8 & 38.1 & 46.7 & 42.5 & 85.4 & 38.4 & \underline{91.8} & 66.4 & 37.0 & 87.8 & 40.7 & 52.4 & \underline{44.6} & 41.7 & \underline{59.0} & 56.1\\
CorDA\cite{wang2021domain} & \underline{94.7} & \underline{63.1} & 87.6 & 30.7 & 40.6 & 40.2 & 47.8 & 51.6 & 87.6 & \underline{47.0} & 89.7 & 66.7 & 35.9 & \underline{90.2} & 48.9 & 57.5 & 0.0 & 39.8 & 56.0 & 56.6\\
ProDA~\cite{zhang2021prototypical} & 87.8 & 56.0 & 79.7 & \underline{46.3} & \underline{44.8} & \underline{45.6} & \underline{53.5} & \underline{53.5} & \underline{88.6} & 45.2 & 82.1 & \underline{70.7} & \underline{39.2} & 88.8 & 45.5 & \underline{59.4} & 1.0 & \underline{48.9} & 56.4 & \underline{57.5}\\
DAFormer (Ours) & \textbf{95.7} & \textbf{70.2} & \textbf{89.4} & \textbf{53.5} & \textbf{48.1} & \textbf{49.6} & \textbf{55.8} & \textbf{59.4} & \textbf{89.9} & \textbf{47.9} & \textbf{92.5} & \textbf{72.2} & \textbf{44.7} & \textbf{92.3} & \textbf{74.5} & \textbf{78.2} & \textbf{65.1} & \textbf{55.9} & \textbf{61.8} & \textbf{68.3}\\
\bottomrule
\end{tabular}
}

\vspace{1cm}
\centering
\caption{Comparison with previous methods on Synthia$\rightarrow$Cityscapes. Our results (DAFormer) are averaged over 3 random seeds.}
\label{tab:sota_synthia}
\setlength{\tabcolsep}{3pt}
\resizebox{\textwidth}{!}{%
\begin{tabular}{l|cccccccccccccccc|cc}
\toprule
 & Road & S.walk & Build. & Wall & Fence & Pole & Tr.Light & Sign & Veget. & Sky & Person & Rider & Car & Bus & M.bike & Bike & mIoU16 & mIoU13\\
\midrule
SPIGAN~\cite{lee2018spigan} & 71.1 & 29.8 & 71.4 & 3.7 & 0.3 & 33.2 & 6.4 & 15.6 & 81.2 & 78.9 & 52.7 & 13.1 & 75.9 & 25.5 & 10.0 & 20.5 & 36.8 & 42.4\\
GIO-Ada~\cite{chen2019learning} & 78.3 & 29.2 & 76.9 & 11.4 & 0.3 & 26.5 & 10.8 & 17.2 & 81.7 & 81.9 & 45.8 & 15.4 & 68.0 & 15.9 & 7.5 & 30.4 & 37.3 & 43.0\\
AdaptSeg~\cite{tsai2018learning} & 79.2 & 37.2 & 78.8 & -- & -- & -- & 9.9 & 10.5 & 78.2 & 80.5 & 53.5 & 19.6 & 67.0 & 29.5 & 21.6 & 31.3 & -- & 45.9\\
PatchAlign~\cite{tsai2019domain} & 82.4 & 38.0 & 78.6 & 8.7 & 0.6 & 26.0 & 3.9 & 11.1 & 75.5 & 84.6 & 53.5 & 21.6 & 71.4 & 32.6 & 19.3 & 31.7 & 40.0 & 46.5\\
CLAN~\cite{luo2019taking} & 81.3 & 37.0 & 80.1 & -- & -- & -- & 16.1 & 13.7 & 78.2 & 81.5 & 53.4 & 21.2 & 73.0 & 32.9 & 22.6 & 30.7 & -- & 47.8\\
ADVENT~\cite{vu2019advent} & 85.6 & 42.2 & 79.7 & 8.7 & 0.4 & 25.9 & 5.4 & 8.1 & 80.4 & 84.1 & 57.9 & 23.8 & 73.3 & 36.4 & 14.2 & 33.0 & 41.2 & 48.0\\
CBST~\cite{zou2018unsupervised} & 68.0 & 29.9 & 76.3 & 10.8 & 1.4 & 33.9 & 22.8 & 29.5 & 77.6 & 78.3 & 60.6 & 28.3 & 81.6 & 23.5 & 18.8 & 39.8 & 42.6 & 48.9\\
DADA~\cite{vu2019dada} & 89.2 & 44.8 & 81.4 & 6.8 & 0.3 & 26.2 & 8.6 & 11.1 & 81.8 & 84.0 & 54.7 & 19.3 & 79.7 & 40.7 & 14.0 & 38.8 & 42.6 & 49.8\\
MRKLD~\cite{zou2019confidence} & 67.7 & 32.2 & 73.9 & 10.7 & 1.6 & 37.4 & 22.2 & 31.2 & 80.8 & 80.5 & 60.8 & 29.1 & 82.8 & 25.0 & 19.4 & 45.3 & 43.8 & 50.1\\
BDL~\cite{li2019bidirectional} & 86.0 & 46.7 & 80.3 & -- & -- & -- & 14.1 & 11.6 & 79.2 & 81.3 & 54.1 & 27.9 & 73.7 & 42.2 & 25.7 & 45.3 & -- & 51.4\\
CAG~\cite{zhang2019category} & 84.7 & 40.8 & 81.7 & 7.8 & 0.0 & 35.1 & 13.3 & 22.7 & 84.5 & 77.6 & 64.2 & 27.8 & 80.9 & 19.7 & 22.7 & 48.3 & 44.5 & 51.5\\
PIT~\cite{lv2020cross} & 83.1 & 27.6 & 81.5 & 8.9 & 0.3 & 21.8 & 26.4 & 33.8 & 76.4 & 78.8 & 64.2 & 27.6 & 79.6 & 31.2 & 31.0 & 31.3 & 44.0 & 51.8\\
SIM~\cite{wang2020differential} & 83.0 & 44.0 & 80.3 & -- & -- & -- & 17.1 & 15.8 & 80.5 & 81.8 & 59.9 & 33.1 & 70.2 & 37.3 & 28.5 & 45.8 & -- & 52.1\\
FDA~\cite{yang2020fda} & 79.3 & 35.0 & 73.2 & -- & -- & -- & 19.9 & 24.0 & 61.7 & 82.6 & 61.4 & 31.1 & 83.9 & 40.8 & 38.4 & 51.1 & -- & 52.5\\
FADA~\cite{wang2020classes} & 84.5 & 40.1 & 83.1 & 4.8 & 0.0 & 34.3 & 20.1 & 27.2 & 84.8 & 84.0 & 53.5 & 22.6 & 85.4 & 43.7 & 26.8 & 27.8 & 45.2 & 52.5\\
APODA~\cite{yang2020adversarial} & 86.4 & 41.3 & 79.3 & -- & -- & -- & 22.6 & 17.3 & 80.3 & 81.6 & 56.9 & 21.0 & 84.1 & 49.1 & 24.6 & 45.7 & -- & 53.1\\
DACS~\cite{tranheden2021dacs} & 80.6 & 25.1 & 81.9 & 21.5 & 2.9 & 37.2 & 22.7 & 24.0 & 83.7 & \textbf{90.8} & 67.6 & 38.3 & 82.9 & 38.9 & 28.5 & 47.6 & 48.3 & 54.8\\
Seg-Uncert.~\cite{zheng2021rectifying} & 87.6 & 41.9 & 83.1 & 14.7 & 1.7 & 36.2 & 31.3 & 19.9 & 81.6 & 80.6 & 63.0 & 21.8 & 86.2 & 40.7 & 23.6 & 53.1 & 47.9 & 54.9\\
CTF~\cite{ma2021coarse} & 75.7 & 30.0 & 81.9 & 11.5 & 2.5 & 35.3 & 18.0 & 32.7 & 86.2 & 90.1 & 65.1 & 33.2 & 83.3 & 36.5 & 35.3 & \underline{54.3} & 48.2 & 55.5\\
IAST~\cite{mei2020instance} & 81.9 & 41.5 & 83.3 & 17.7 & 4.6 & 32.3 & 30.9 & 28.8 & 83.4 & 85.0 & 65.5 & 30.8 & 86.5 & 38.2 & 33.1 & 52.7 & 49.8 & 57.0\\
SAC~\cite{araslanov2021self} & \underline{89.3} & \underline{47.2} & \underline{85.5} & 26.5 & 1.3 & 43.0 & 45.5 & 32.0 & \underline{87.1} & 89.3 & 63.6 & 25.4 & 86.9 & 35.6 & 30.4 & 53.0 & 52.6 & 59.3\\
ProDA~\cite{zhang2021prototypical} & 87.8 & 45.7 & 84.6 & \underline{37.1} & 0.6 & \underline{44.0} & \underline{54.6} & 37.0 & \textbf{88.1} & 84.4 & \textbf{74.2} & 24.3 & \textbf{88.2} & \underline{51.1} & \underline{40.5} & 45.6 & \underline{55.5} & 62.0\\
CorDA~\cite{wang2021domain} & \textbf{93.3} & \textbf{61.6} & 85.3 & 19.6 & \underline{5.1} & 37.8 & 36.6 & \underline{42.8} & 84.9 & \underline{90.4} & 69.7 & \underline{41.8} & 85.6 & 38.4 & 32.6 & 53.9 & 55.0 & \underline{62.8}\\
DAFormer (Ours) & 84.5 & 40.7 & \textbf{88.4} & \textbf{41.5} & \textbf{6.5} & \textbf{50.0} & \textbf{55.0} & \textbf{54.6} & 86.0 & 89.8 & \underline{73.2} & \textbf{48.2} & \underline{87.2} & \textbf{53.2} & \textbf{53.9} & \textbf{61.7} & \textbf{60.9} & \textbf{67.4}\\
\bottomrule
\end{tabular}
}
\vspace{1.4cm}
\end{table*}

In the main paper, we compare DAFormer with a selection of representative UDA methods. However, various other UDA methods were proposed in the last few years. A comprehensive comparison with these is shown in Tab.~\ref{tab:sota_gta} for GTA$\rightarrow$Cityscapes and in Tab.~\ref{tab:sota_synthia} for Synthia$\rightarrow$Cityscapes. On the one side, some of the newly shown methods can achieve a higher IoU for specific classes than the previous methods shown in the main paper, but on the other side, their performance suffers for other classes. Therefore, ProDA~\cite{zhang2021prototypical} still achieves the best mIoU of the previous state-of-the-art methods. Overall, DAFormer is able to outperform all previous works both in mIoU and classwise IoU for GTA$\rightarrow$Cityscapes, often by a considerable margin. On Synthia$\rightarrow$Cityscapes, this statement holds except for the stuff-classes \emph{road}, \emph{sidewalk}, \emph{vegetation}, and \emph{sky}. This might be due to the shape-bias of Transformers~\cite{bhojanapalli2021understanding}, which causes the network to focus more on shape than texture. The shape bias could improve the generalization ability for thing-classes as their shape is more domain-robust than their texture. However, for stuff-classes, the texture is sometimes crucial to distinguish similar classes such as \emph{road} and \emph{sidewalk} and a shape-bias could be hindering.

The results in Tab.~\ref{tab:sota_gta} and Tab.~\ref{tab:sota_synthia} are reported with the training configurations used in the original methods, which do not use learning rate warmup. For a fair comparison, we have re-implemented DACS~\cite{tranheden2021dacs} with our training configuration including learning rate warmup, which achieves 54.2 mIoU on GTA$\rightarrow$Cityscapes. Still, DAFormer outperforms it by +14.1 mIoU.

DAFormer uses the last checkpoint of a training for evaluation. The reported results are averaged over three training runs and have a standard deviation of 0.5 mIoU on both benchmarks, which shows that the training process is stable. We do not use the target validation dataset for checkpoint selection in contrast to some other works~\cite{zhang2019category,zhang2021prototypical}.

\section{Discussion}
\label{sec:suppl_discussion}

\subsection{Limitations}

Due to computational constraints, we only use a selection of network architectures to support our claims. However, there are further interesting architectures that could be explored in the future such as~\cite{yuan2021ocnet, liu2021swin}. Also, other training aspects such as larger batch sizes could be of relevance.

In Fig. 3 of the main paper and in the discussion of the Synthia results in Sec.~\ref{sec:sota}, there have been some indications that a Transformer architecture is not ideal for stuff-classes. This might be due to the shape-bias of Transformers~\cite{bhojanapalli2021understanding}. The focus on shape instead of texture might be disadvantageous for the distinction of stuff-classes as the texture is an important aspect for their recognition. Even though a further investigation is out of the scope of this work, we believe that this is an interesting aspect for future work, which could potentially lead to a network architecture, even better suited for UDA.

Context-aware fusion assumes that context correlations are domain-invariant. This is often the case for the typical UDA benchmarks~\cite{zhou2021context}. However, this assumption can break down for some special cases in other domains, where the context misleads the model (e.g. misclassification of a cow on road as a horse)~\cite{hoyer2019grid}.

RCS is designed to counter a long-tail data distribution of the source dataset.
It is unproblematic for RCS if a class is more common in the target dataset than in the source dataset (e.g. bicycle) as it is balanced by RCS on the source domain and regularly sampled on the target domain. If a class is extremely rare in the target dataset, it might happen that this class is not sampled often enough for efficient adaptation. Therefore, the pseudo-labels would not contain this class and, conceptually, it would not be possible to specifically select samples with this class from the target data.

As shown in the error cases in Sec.~\ref{sec:qualitative_examples}, our method struggles with differences in the annotation policy between source training data and target evaluation data. One example is the bicycle wheel. While the entire wheel is segmented in Cityscapes (see Fig.~\ref{fig:label_policy_bicycle_cityscapes}), only the tires and spokes are segmented in GTA (see Fig.~\ref{fig:label_policy_bicycle_gta}). Also, there are corner cases, where the annotation policy is not defined by source labels such as cycle paths on road (see the top row in Fig.~\ref{fig:error_cases_wall}) or small busses (see the bottom row in Fig.~\ref{fig:error_cases_vans}).
In order to resolve these issues, additional information about the annotations policy and corner cases would be necessary. A potential solution might be the use of a few target training labels as studied in semi-supervised domain adaptation.

\subsection{Potential Negative Impact}
Our work improves the adaptability of semantic segmentation, which can be used to enable many good applications such as autonomous driving. However, UDA might also be utilized in undesired applications such as surveillance or military UAVs. This is a general problem of improving semantic segmentation algorithms. A possible counter-measure could be legal restrictions of the use cases for semantic segmentation algorithms.

%% file: preds/prediction_head.tex
{\footnotesize
\begin{tabularx}{\linewidth}{*{6}{Y}}
Image & DeepLabV2 Src-Only & SegFormer Src-Only & ProDA~\cite{zhang2021prototypical} & DAFormer (Ours) & Ground Truth \\
\end{tabularx}
} %

%% file: preds/8.tex
\begin{tikzpicture}

	\draw (0.0, 0.0) node[inner sep=0pt] (image) {\includegraphics[width=\linewidth]{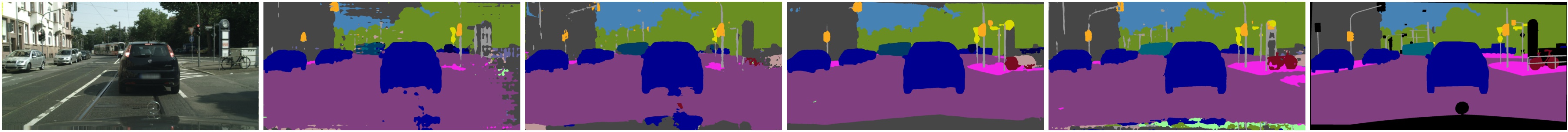}};

    \draw (4.2, 0.2) node[draw=white, rectangle, densely dotted, very thick, fill opacity=0.0, minimum width=0.25in, minimum height=0.15in] (rect) {};

\end{tikzpicture}

%% file: preds/91.tex
\begin{tikzpicture}

	\draw (0.0, 0.0) node[inner sep=0pt] (image) {\includegraphics[width=\linewidth]{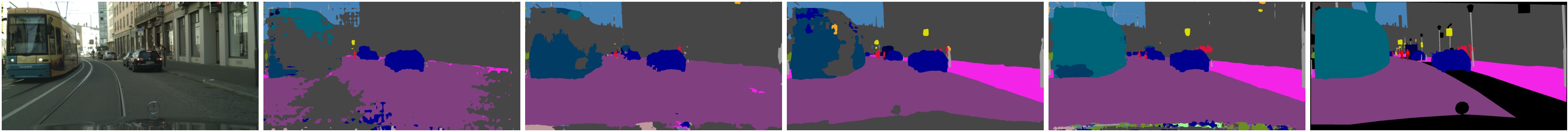}};

    \draw (3.4, 0.25) node[draw=white, rectangle, densely dotted, very thick, fill opacity=0.0, minimum width=0.35in, minimum height=0.35in] (rect) {};

\end{tikzpicture}

%% file: preds/116.tex
\begin{tikzpicture}

	\draw (0.0, 0.0) node[inner sep=0pt] (image) {\includegraphics[width=\linewidth]{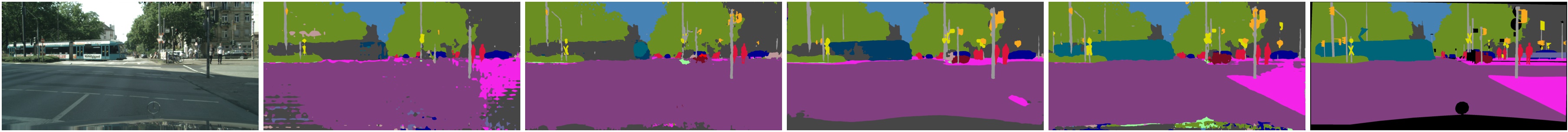}};

    \draw (3.7, 0.2) node[draw=white, rectangle, densely dotted, very thick, fill opacity=0.0, minimum width=0.55in, minimum height=0.15in] (rect) {};

\end{tikzpicture}

%% file: preds/228.tex
\begin{tikzpicture}
	\draw (0.0, 0.0) node[inner sep=0pt] (image) {\includegraphics[width=\linewidth]{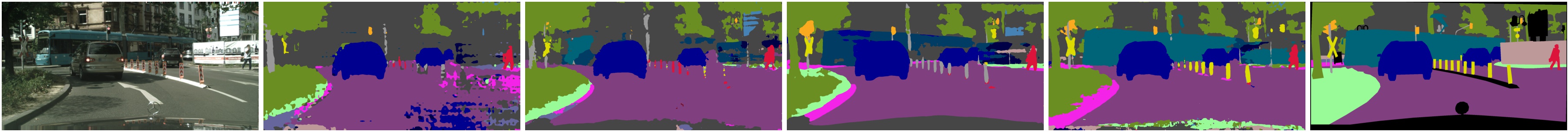}};
    \draw (4.2, 0.2) node[draw=white, rectangle, densely dotted, very thick, fill opacity=0.0, minimum width=0.8in, minimum height=0.25in] (rect) {};
\end{tikzpicture}

%% file: preds/palette.tex
\scriptsize
\setlength\tabcolsep{1pt}
{
\newcolumntype{P}[1]{>{\centering\arraybackslash}p{#1}}
\begin{tabular}{@{}*{20}{P{0.09\columnwidth}}@{}}
     {\cellcolor[rgb]{0.5,0.25,0.5}}\textcolor{white}{road} 
     &{\cellcolor[rgb]{0.957,0.137,0.91}}sidew. 
     &{\cellcolor[rgb]{0.275,0.275,0.275}}\textcolor{white}{build.} 
     &{\cellcolor[rgb]{0.4,0.4,0.612}}\textcolor{white}{wall} 
     &{\cellcolor[rgb]{0.745,0.6,0.6}}fence 
     &{\cellcolor[rgb]{0.6,0.6,0.6}}pole 
     &{\cellcolor[rgb]{0.98,0.667,0.118}}tr. light
     &{\cellcolor[rgb]{0.863,0.863,0}}tr. sign 
     &{\cellcolor[rgb]{0.42,0.557,0.137}}veget. 
     &{\cellcolor[rgb]{0.596,0.984,0.596}}terrain 
     &{\cellcolor[rgb]{0.275,0.510,0.706}}sky
     &{\cellcolor[rgb]{0.863,0.078,0.235}}\textcolor{white}{person} 
     &{\cellcolor[rgb]{1,0,0}}\textcolor{white}{rider} 
     &{\cellcolor[rgb]{0,0,0.557}}\textcolor{white}{car} 
     &{\cellcolor[rgb]{0,0,0.275}}\textcolor{white}{truck} 
     &{\cellcolor[rgb]{0,0.235,0.392}}\textcolor{white}{bus}
     &{\cellcolor[rgb]{0,0.392,0.471}}\textcolor{white}{train} 
     &{\cellcolor[rgb]{0,0,0.902}}\textcolor{white}{m.bike} 
     & {\cellcolor[rgb]{0.467,0.043,0.125}}\textcolor{white}{bike}
     &{\cellcolor[rgb]{0,0,0}}\textcolor{white}{n/a.}
\end{tabular}
}

%% file: preds/51.tex
\begin{tikzpicture}

	\draw (0.0, 0.0) node[inner sep=0pt] (image) {\includegraphics[width=\linewidth]{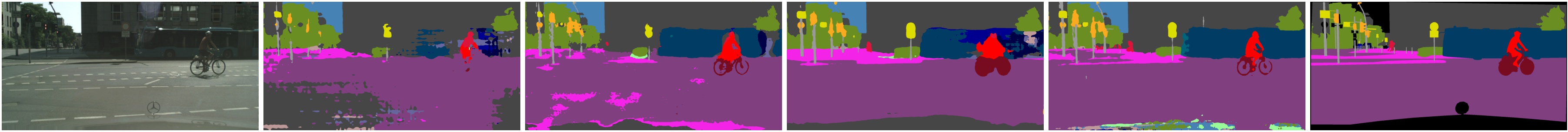}};

    \draw (5.05, 0.25) node[draw=white, rectangle, densely dotted, very thick, fill opacity=0.0, minimum width=0.55in, minimum height=0.2in] (rect) {};

\end{tikzpicture}

%% file: preds/139.tex
\begin{tikzpicture}

	\draw (0.0, 0.0) node[inner sep=0pt] (image) {\includegraphics[width=\linewidth]{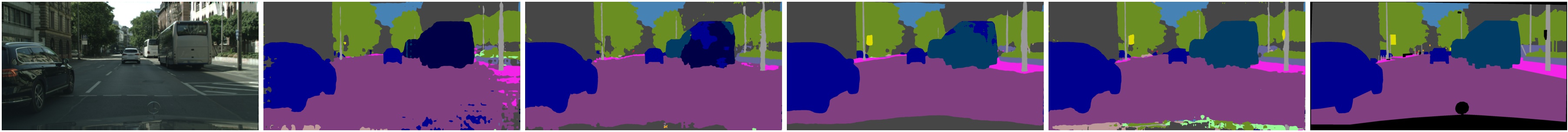}};

    \draw (4.95, 0.25) node[draw=white, rectangle, densely dotted, very thick, fill opacity=0.0, minimum width=0.3in, minimum height=0.25in] (rect) {};

\end{tikzpicture}

%% file: preds/200.tex
\begin{tikzpicture}

	\draw (0.0, 0.0) node[inner sep=0pt] (image) {\includegraphics[width=\linewidth]{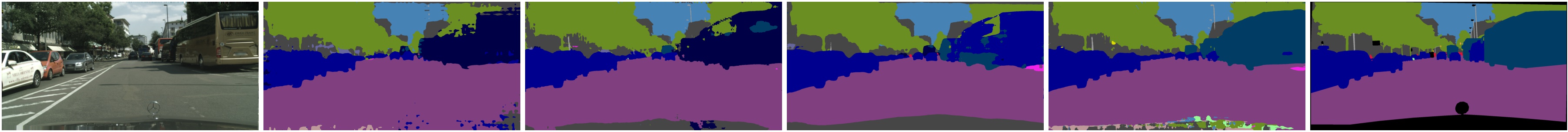}};

    \draw (5.15, 0.25) node[draw=white, rectangle, densely dotted, very thick, fill opacity=0.0, minimum width=0.5in, minimum height=0.35in] (rect) {};

\end{tikzpicture}%

%% file: preds/454.tex
\begin{tikzpicture}

	\draw (0.0, 0.0) node[inner sep=0pt] (image) {\includegraphics[width=\linewidth]{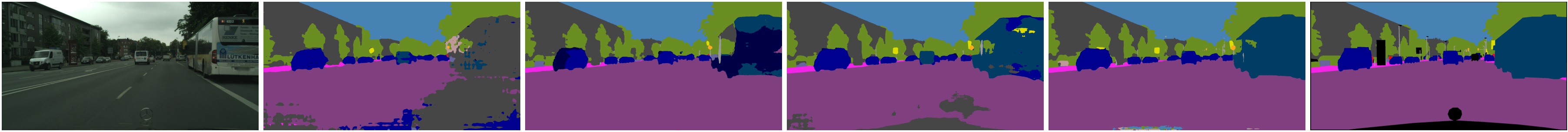}};

    \draw (5.35, 0.2) node[draw=white, rectangle, densely dotted, very thick, fill opacity=0.0, minimum width=0.35in, minimum height=0.32in] (rect) {};

\end{tikzpicture}

%% file: preds/21.tex
\begin{tikzpicture}

	\draw (0.0, 0.0) node[inner sep=0pt] (image) {\includegraphics[width=\linewidth]{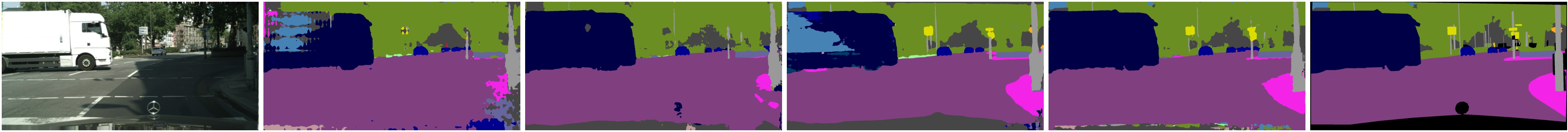}};

    \draw (3.65, 0.25) node[draw=white, rectangle, densely dotted, very thick, fill opacity=0.0, minimum width=0.52in, minimum height=0.35in] (rect) {};

\end{tikzpicture}

%% file: preds/134.tex
\begin{tikzpicture}

	\draw (0.0, 0.0) node[inner sep=0pt] (image) {\includegraphics[width=\linewidth]{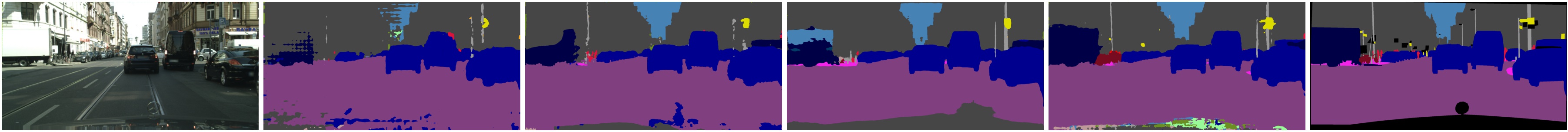}};

    \draw (3.3, 0.25) node[draw=white, rectangle, densely dotted, very thick, fill opacity=0.0, minimum width=0.25in, minimum height=0.25in] (rect) {};

\end{tikzpicture}

%% file: preds/268.tex
\begin{tikzpicture}

	\draw (0.0, 0.0) node[inner sep=0pt] (image) {\includegraphics[width=\linewidth]{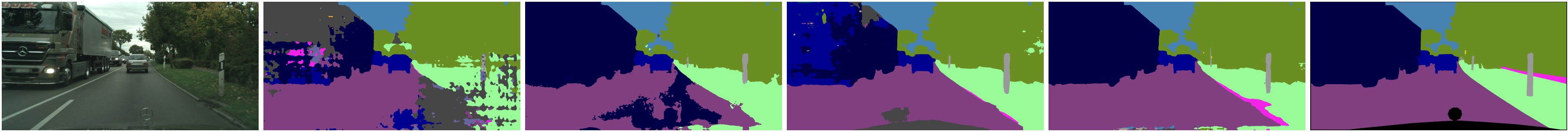}};

    \draw (3.6, 0.2) node[draw=white, rectangle, densely dotted, very thick, fill opacity=0.0, minimum width=0.5in, minimum height=0.4in] (rect) {};

\end{tikzpicture}

%% file: preds/429.tex
\begin{tikzpicture}

	\draw (0.0, 0.0) node[inner sep=0pt] (image) {\includegraphics[width=\linewidth]{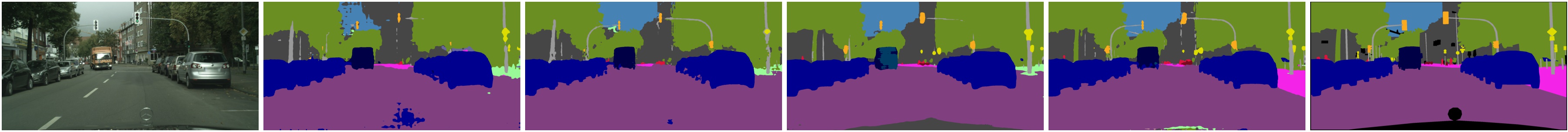}};

    \draw (4.05, 0.1) node[draw=white, rectangle, densely dotted, very thick, fill opacity=0.0, minimum width=0.15in, minimum height=0.15in] (rect) {};

\end{tikzpicture}

%% file: preds/76.tex
\begin{tikzpicture}

	\draw (0.0, 0.0) node[inner sep=0pt] (image) {\includegraphics[width=\linewidth]{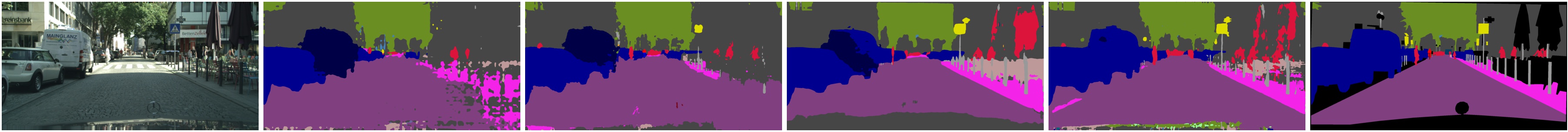}};

    \draw (3.7, 0.15) node[draw=white, rectangle, densely dotted, very thick, fill opacity=0.0, minimum width=0.3in, minimum height=0.25in] (rect) {};

\end{tikzpicture}

%% file: preds/133.tex
\begin{tikzpicture}

	\draw (0.0, 0.0) node[inner sep=0pt] (image) {\includegraphics[width=\linewidth]{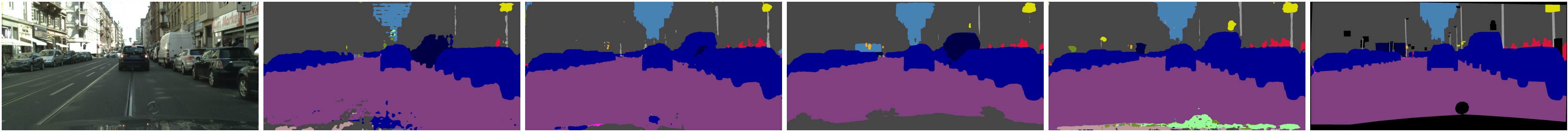}};

    \draw (4.9, 0.2) node[draw=white, rectangle, densely dotted, very thick, fill opacity=0.0, minimum width=0.2in, minimum height=0.2in] (rect) {};

\end{tikzpicture}

%% file: preds/155.tex
\begin{tikzpicture}

	\draw (0.0, 0.0) node[inner sep=0pt] (image) {\includegraphics[width=\linewidth]{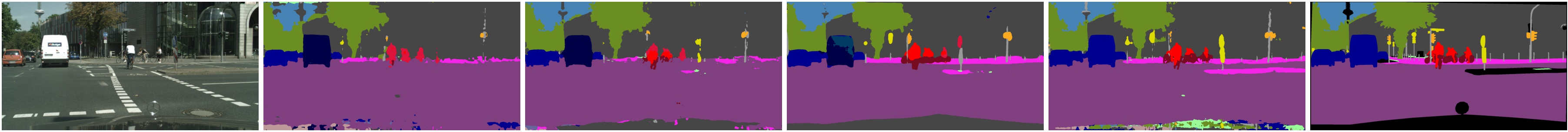}};

    \draw (3.55, 0.2) node[draw=white, rectangle, densely dotted, very thick, fill opacity=0.0, minimum width=0.2in, minimum height=0.2in] (rect) {};

\end{tikzpicture}%

%% file: preds/364.tex
\begin{tikzpicture}

	\draw (0.0, 0.0) node[inner sep=0pt] (image) {\includegraphics[width=\linewidth]{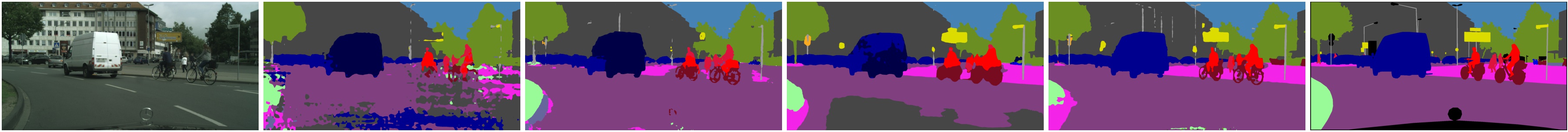}};

    \draw (3.9, 0.15) node[draw=white, rectangle, densely dotted, very thick, fill opacity=0.0, minimum width=0.35in, minimum height=0.25in] (rect) {};

\end{tikzpicture}

%% file: preds/23.tex
\begin{tikzpicture}

	\draw (0.0, 0.0) node[inner sep=0pt] (image) {\includegraphics[width=\linewidth]{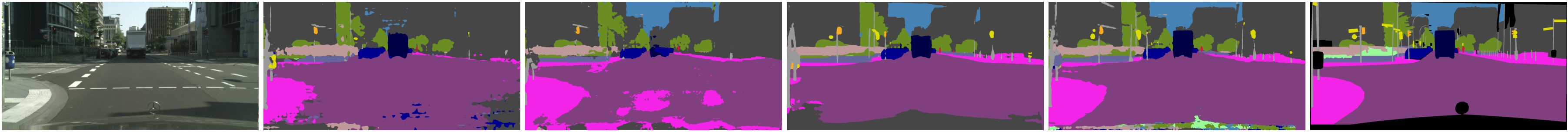}};

    \draw (3.35, -0.3) node[draw=white, rectangle, densely dotted, very thick, fill opacity=0.0, minimum width=0.3in, minimum height=0.3in] (rect) {};

\end{tikzpicture}

%% file: preds/80.tex
\begin{tikzpicture}

	\draw (0.0, 0.0) node[inner sep=0pt] (image) {\includegraphics[width=\linewidth]{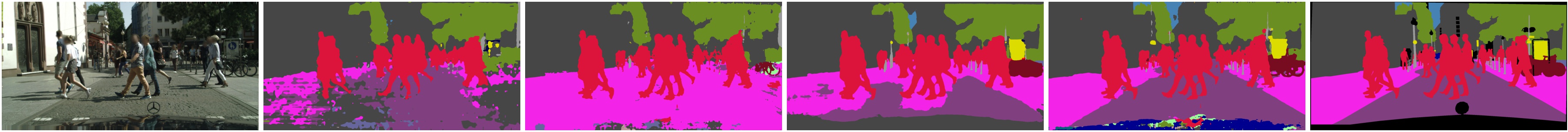}};

    \draw (4.2, -0.25) node[draw=white, rectangle, densely dotted, very thick, fill opacity=0.0, minimum width=0.6in, minimum height=0.2in] (rect) {};

\end{tikzpicture}

%% file: preds/151.tex
\begin{tikzpicture}

	\draw (0.0, 0.0) node[inner sep=0pt] (image) {\includegraphics[width=\linewidth]{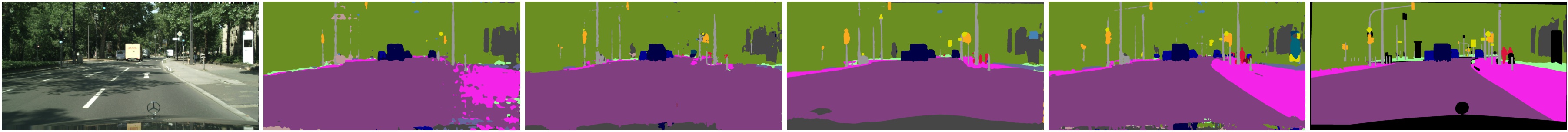}};

    \draw (5.35, -0.2) node[draw=white, rectangle, densely dotted, very thick, fill opacity=0.0, minimum width=0.35in, minimum height=0.32in] (rect) {};

\end{tikzpicture}%

%% file: preds/349.tex
\begin{tikzpicture}

	\draw (0.0, 0.0) node[inner sep=0pt] (image) {\includegraphics[width=\linewidth]{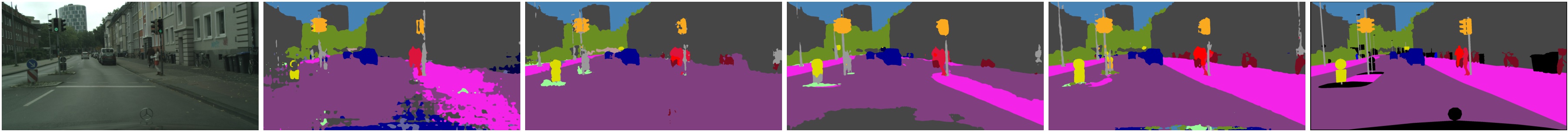}};

    \draw (5.35, -0.1) node[draw=white, rectangle, densely dotted, very thick, fill opacity=0.0, minimum width=0.35in, minimum height=0.25in] (rect) {};

\end{tikzpicture}

%% file: preds/424.tex
\begin{tikzpicture}

	\draw (0.0, 0.0) node[inner sep=0pt] (image) {\includegraphics[width=\linewidth]{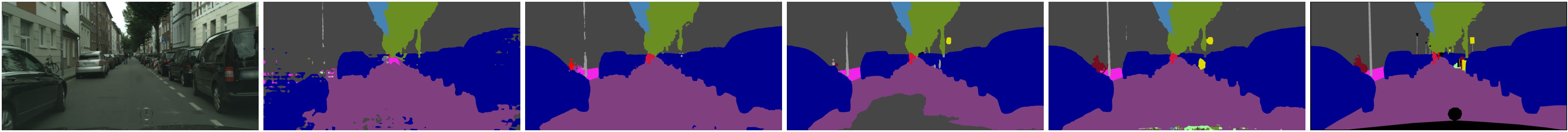}};

    \draw (3.5, 0.0) node[draw=white, rectangle, densely dotted, very thick, fill opacity=0.0, minimum width=0.15in, minimum height=0.15in] (rect) {};
\end{tikzpicture}

%% file: preds/90.tex
\begin{tikzpicture}

	\draw (0.0, 0.0) node[inner sep=0pt] (image) {\includegraphics[width=\linewidth]{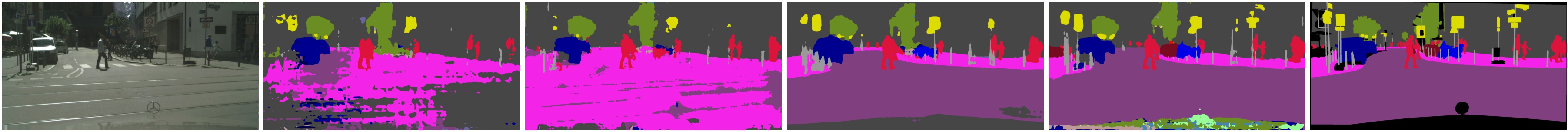}};

    \draw (4.3, 0.18) node[draw=white, rectangle, densely dotted, very thick, fill opacity=0.0, minimum width=0.1in, minimum height=0.1in] (rect) {};
\end{tikzpicture}

%% file: preds/366.tex
\begin{tikzpicture}

	\draw (0.0, 0.0) node[inner sep=0pt] (image) {\includegraphics[width=\linewidth]{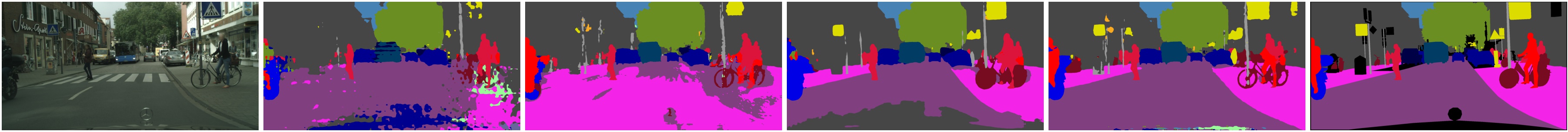}};

    \draw (3.2, -0.05) node[draw=white, rectangle, densely dotted, very thick, fill opacity=0.0, minimum width=0.2in, minimum height=0.35in] (rect) {};
\end{tikzpicture}

%% file: preds/455.tex
\begin{tikzpicture}

	\draw (0.0, 0.0) node[inner sep=0pt] (image) {\includegraphics[width=\linewidth]{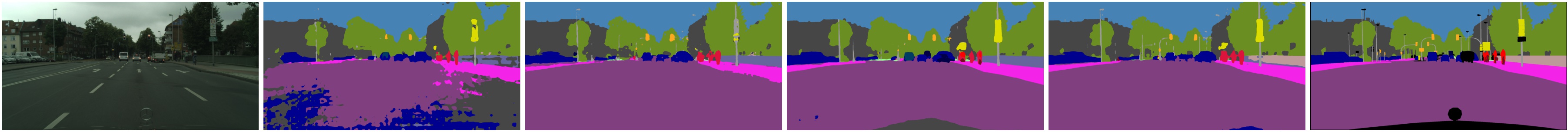}};

    \draw (5.45, 0.05) node[draw=white, rectangle, densely dotted, very thick, fill opacity=0.0, minimum width=0.25in, minimum height=0.1in] (rect) {};
\end{tikzpicture}

%% file: preds/19.tex
\begin{tikzpicture}

	\draw (0.0, 0.0) node[inner sep=0pt] (image) {\includegraphics[width=\linewidth]{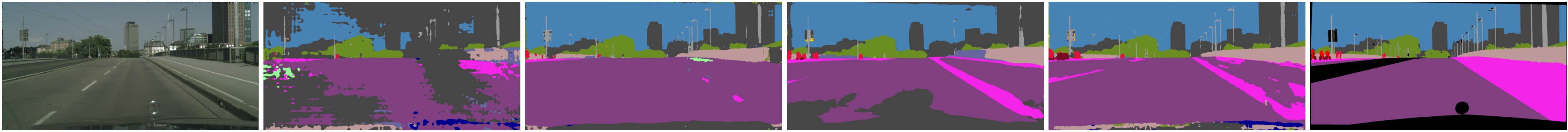}};

    \draw (5.35, -0.15) node[draw=white, rectangle, densely dotted, very thick, fill opacity=0.0, minimum width=0.35in, minimum height=0.2in] (rect) {};

\end{tikzpicture}

%% file: preds/100.tex
\begin{tikzpicture}

	\draw (0.0, 0.0) node[inner sep=0pt] (image) {\includegraphics[width=\linewidth]{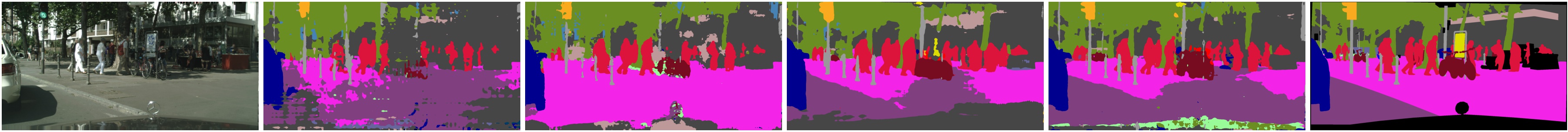}};

    \draw (4.9, -0.3) node[draw=white, rectangle, densely dotted, very thick, fill opacity=0.0, minimum width=0.6in, minimum height=0.2in] (rect) {};

\end{tikzpicture}

%% file: preds/278.tex
\begin{tikzpicture}

	\draw (0.0, 0.0) node[inner sep=0pt] (image) {\includegraphics[width=\linewidth]{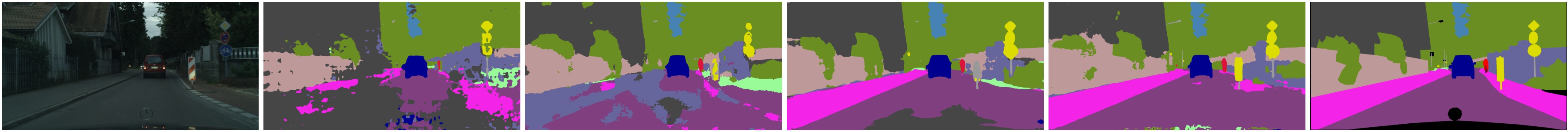}};

    \draw (5.35, -0.4) node[draw=white, rectangle, densely dotted, very thick, fill opacity=0.0, minimum width=0.35in, minimum height=0.2in] (rect) {};
    
    \draw (5.28, 0.13) node[draw=white, rectangle, densely dotted, very thick, fill opacity=0.0, minimum width=0.4in, minimum height=0.1in] (rect) {};

\end{tikzpicture}

%% file: preds/257.tex
\begin{tikzpicture}

	\draw (0.0, 0.0) node[inner sep=0pt] (image) {\includegraphics[width=\linewidth]{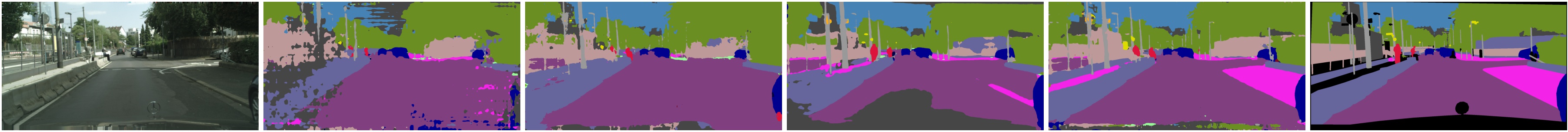}};

    \draw (5.2, 0.2) node[draw=white, rectangle, densely dotted, very thick, fill opacity=0.0, minimum width=0.4in, minimum height=0.15in] (rect) {};

\end{tikzpicture}%

%% file: preds/438.tex
\begin{tikzpicture}

	\draw (0.0, 0.0) node[inner sep=0pt] (image) {\includegraphics[width=\linewidth]{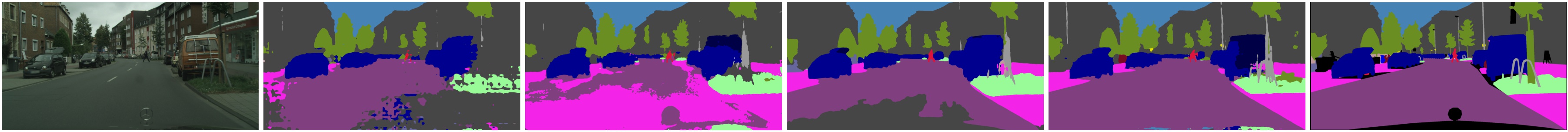}};

    \draw (5.15, 0.1) node[draw=white, rectangle, densely dotted, very thick, fill opacity=0.0, minimum width=0.25in, minimum height=0.25in] (rect) {};

\end{tikzpicture}

%% file: preds/201.tex
\begin{tikzpicture}

	\draw (0.0, 0.0) node[inner sep=0pt] (image) {\includegraphics[width=\linewidth]{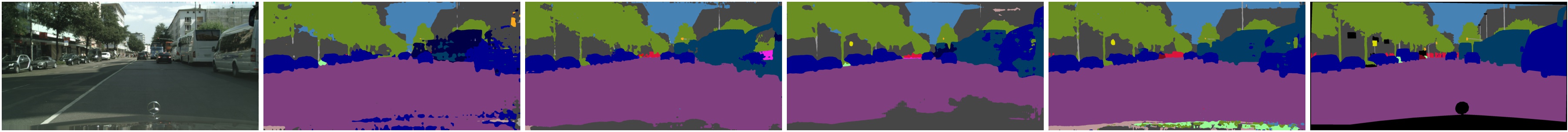}};

    \draw (5.5, 0.18) node[draw=white, rectangle, densely dotted, very thick, fill opacity=0.0, minimum width=0.25in, minimum height=0.25in] (rect) {};

\end{tikzpicture}%

%% file: preds/71.tex
\begin{tikzpicture}

	\draw (0.0, 0.0) node[inner sep=0pt] (image) {\includegraphics[width=\linewidth]{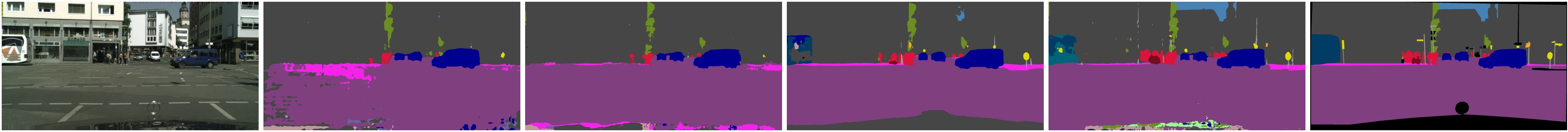}};

    \draw (3.15, 0.2) node[draw=white, rectangle, densely dotted, very thick, fill opacity=0.0, minimum width=0.15in, minimum height=0.2in] (rect) {};

\end{tikzpicture}%

%% file: preds/357.tex
\begin{tikzpicture}

	\draw (0.0, 0.0) node[inner sep=0pt] (image) {\includegraphics[width=\linewidth]{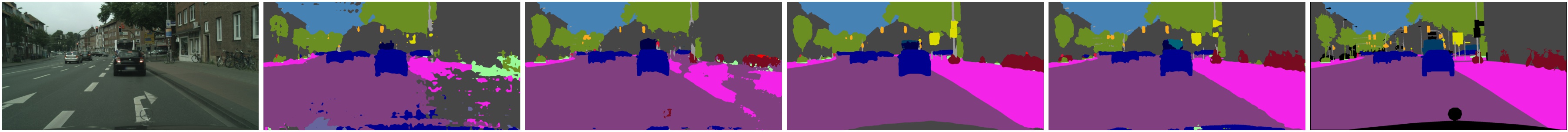}};

    \draw (4.35, 0.2) node[draw=white, rectangle, densely dotted, very thick, fill opacity=0.0, minimum width=0.15in, minimum height=0.15in] (rect) {};

\end{tikzpicture}

%% file: preds/29.tex
\begin{tikzpicture}

	\draw (0.0, 0.0) node[inner sep=0pt] (image) {\includegraphics[width=\linewidth]{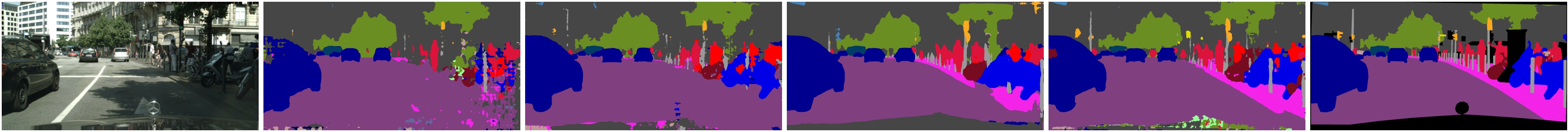}};

    \draw (5.35, 0.15) node[draw=white, rectangle, densely dotted, very thick, fill opacity=0.0, minimum width=0.35in, minimum height=0.15in] (rect) {};

\end{tikzpicture}%

%% file: preds/38.tex
\begin{tikzpicture}

	\draw (0.0, 0.0) node[inner sep=0pt] (image) {\includegraphics[width=\linewidth]{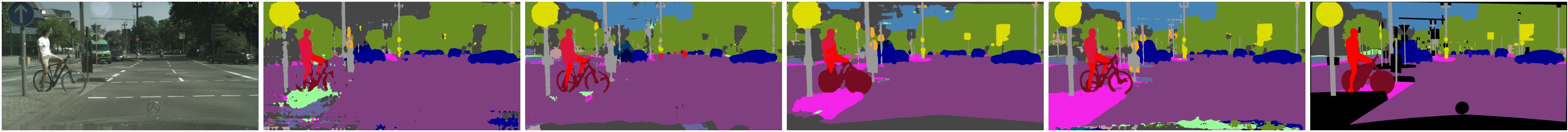}};

    \draw (3.45, 0.25) node[draw=white, rectangle, densely dotted, very thick, fill opacity=0.0, minimum width=0.15in, minimum height=0.15in] (rect) {};
    
    \draw (3.6, -0.2) node[draw=white, rectangle, densely dotted, very thick, fill opacity=0.0, minimum width=0.3in, minimum height=0.15in] (rect) {};

\end{tikzpicture}%